\documentclass{article}

\usepackage{microtype}
\usepackage{graphicx}
\usepackage{subcaption}
\usepackage{booktabs} 

\usepackage[hyphens]{url}
\usepackage{hyperref}

\usepackage[accepted]{icml2026}
\usepackage{amsmath}
\usepackage{amssymb}
\usepackage{mathtools}
\usepackage{amsthm}
\usepackage{svg}
\usepackage{graphicx}
\usepackage{xcolor}
\usepackage{listings}
\usepackage[most]{tcolorbox}
\tcbuselibrary{listings,skins,breakable}
\usepackage{placeins}
\newtcolorbox{databox}[2][]{%
  float*=htb,
  floatplacement=htb,
  width=\textwidth,
  before skip=1em, after skip=1em,
  enhanced, breakable,
  colback=gray!5!white,
  colframe=gray!75!black,
  colbacktitle=gray!75!black,
  coltitle=white,
  fonttitle=\bfseries,
  title={#2},
  listing engine=listings,
  listing only,
  listing options={
    basicstyle=\ttfamily\small,
    breaklines=true,
    breakanywhere=true,
  },
  #1
}

\usepackage[capitalize,noabbrev]{cleveref}

\theoremstyle{plain}
\newtheorem{theorem}{Theorem}[section]

\theoremstyle{definition}
\newtheorem{definition}[theorem]{Definition}

\theoremstyle{remark}

\usepackage[disable,textsize=tiny]{todonotes}
\icmltitlerunning{Dynamics Within Latent Chain-of-Thought: An Empirical Study of Causal Structure}

\begin{document}

\twocolumn[
  \icmltitle{Dynamics Within Latent Chain-of-Thought: \\ An Empirical Study of Causal Structure}



  \icmlsetsymbol{equal}{*}

\begin{icmlauthorlist}
  \icmlauthor{Zirui Li}{hitsz}
  \icmlauthor{Xuefeng Bai}{hitsz}
  \icmlauthor{Kehai Chen}{hitsz}
  \icmlauthor{Yizhi Li}{map,manchester}
  \icmlauthor{Jian Yang}{map,beihang}
  \icmlauthor{Chenghua Lin}{manchester}
  \icmlauthor{Min Zhang}{hitsz}
\end{icmlauthorlist}

\icmlaffiliation{hitsz}{Harbin Institute of Technology, Shenzhen, China}
\icmlaffiliation{map}{M-A-P}
\icmlaffiliation{manchester}{University of Manchester, United Kingdom}
\icmlaffiliation{beihang}{Beihang University, China}

\icmlcorrespondingauthor{Xuefeng Bai}{baixuefeng@hit.edu.cn}

\icmlkeywords{Latent Chain-of-Thought, Causal Intervention, Representation Analysis, Reasoning}


  \vskip 0.3in
]



\printAffiliationsAndNotice{}  

\begin{abstract}
Latent or continuous chain-of-thought methods replace explicit textual rationales with a number of internal latent steps, but these intermediate computations are difficult to evaluate beyond correlation-based probes. In this paper, we view latent chain-of-thought as a manipulable causal process in representation space by modeling latent steps as variables in a structural causal model (SCM) and analyzing their effects through step-wise $\mathrm{do}$-interventions. 
We study two representative paradigms (i.e., Coconut and CODI) on both mathematical and general reasoning tasks to investigate three key questions: (1) which steps are causally necessary for correctness and when answers become decodable early;
(2) how influence propagates across steps and how this structure compares to explicit CoT; and
(3) whether intermediate trajectories retain competing answer modes and how output-level commitment differs from representational commitment across steps. We find that latent-step budgets behave less like homogeneous extra depth and more like staged functionality with non-local routing, and we identify a persistent gap between early output bias and late representational commitment. These results motivate mode-conditional and stability-aware analyses, together with corresponding training/decoding objectives, as more reliable tools for interpreting and improving latent reasoning systems. Code is available at \url{https://github.com/J1mL1/causal-latent-cot}.
\end{abstract}

\section{Introduction}

Large language models (LLMs) have achieved strong performance on mathematical problem solving and logical question answering \citep{cobbe_training_2021, geva-etal-2021-aristotle}. A widely adopted technique is Chain-of-Thought (CoT) prompting, which improves accuracy by eliciting intermediate reasoning steps in natural language \citep{wei_chain--thought_2022}. Despite its empirical effectiveness, explicit CoT incurs substantial decoding cost, often produces verbose outputs, and may contain post-hoc rationalizations that do not faithfully reflect the computations driving model predictions \citep{pruthi_learning_2020, turpin_language_2023}. These limitations motivate a shift from reasoning in tokens to reasoning in representations.

Recent work explores latent/continuous CoT, where multi-step inference is carried out in continuous hidden representations rather than long textual explanations \citep{hao_training_2024, shen_codi_2025, zhang_soft_2025, xu_softcot_2025, wei_sim-cot_2025, gozeten_continuous_2025}. This paradigm promises a higher-bandwidth internal workspace and reduced decoding overhead, but it faces a fundamental interpretability challenge: 1) intermediate computations are no longer exposed as discrete, human-editable steps; 2) reasoning-relevant information is often distributed across latent dimensions and iterative steps. 
Hence, traditional analytical methods—such as step editing or ablation—cannot be directly applied to implicit CoT, leaving the causal and mechanistic analysis of implicit CoT underexplored.

\begin{figure*}[ht]
    \centering
    \includegraphics[width=\linewidth]{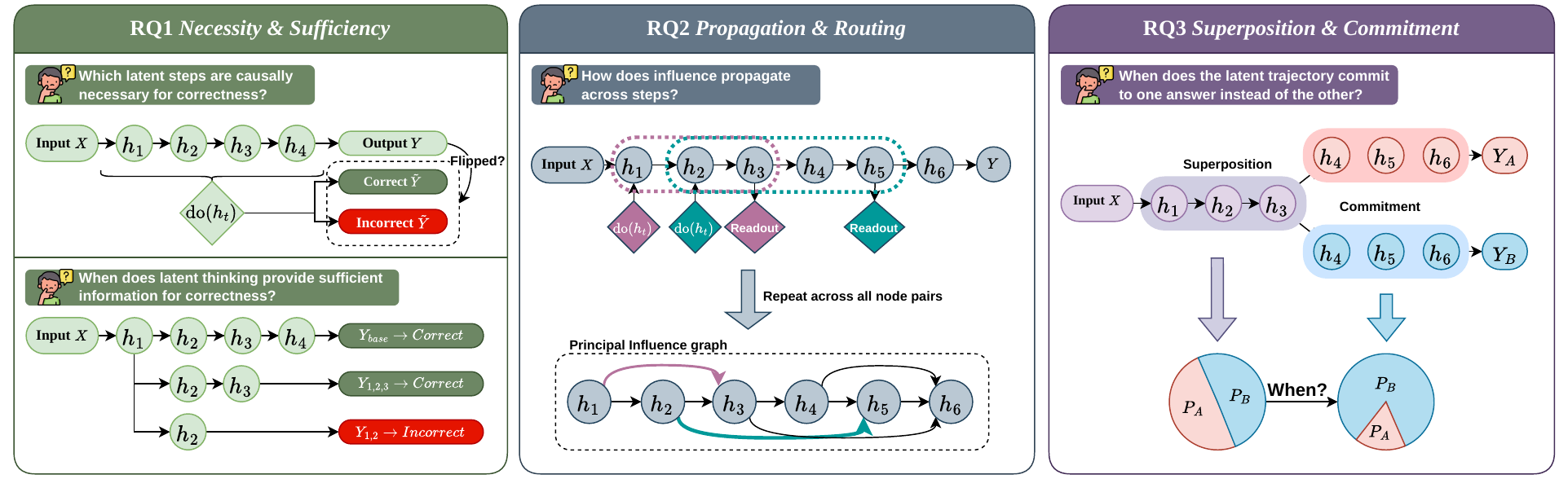}
    \caption{\textbf{Overview of step-centric research questions for latent CoT.}
    RQ1 tests step necessity and early decodability; RQ2 characterizes step-to-step influence propagation; RQ3 probes trajectory-level superposition and commitment across rollouts.}
    \label{fig:rq_overview}
    \vspace{-0.3cm}
\end{figure*}

Motivated by this research gap, we conceptualize latent CoT as a \textbf{causal system} \citep{pearl_causality_2022,causal_survey} evolving over latent-step variables and evaluate it with intervention-based causal analysis \citep{robins_identifiability_1992,imai_general_2010}.
Specifically, we treat model-defined intermediate latent states as the unit of a reasoning ``step'' and perform step-wise $\mathrm{do}$-interventions \citep{singh_kernel_2022,kaddour_causal_2022} that modify an intermediate state while keeping downstream computation unchanged. We then quantify the causal sensitivity of the model's output to these interventions. Furthermore, we aggregate step-to-step influences to construct a directed acyclic graph \citep{dag} of information flow, derived from intervention-induced shifts in teacher-forced output distributions. This graph captures how information propagates through the latent reasoning process.
Figure~\ref{fig:rq_overview} provides a high-level roadmap of the three questions we study: \textbf{(RQ1)} which latent steps are causally necessary for correctness and when the answer becomes decodable, \textbf{(RQ2)} how influence propagates across steps, and \textbf{(RQ3)} whether intermediate trajectories retain competing hypotheses and how commitment evolves across steps.

We instantiate this causal evaluation framework on two representative latent-reasoning paradigms, Coconut \citep{hao_training_2024} and CODI \citep{shen_codi_2025}, across both mathematical and general reasoning tasks. Empirically, we find a phenomenon--mechanism--nature pattern: \textbf{\emph{(phenomenon)}} causal leverage is highly heterogeneous across latent steps, with a small subset exerting outsized influence; \textbf{\emph{(mechanism)}} step-to-step effects are often non-local, indicating routed propagation rather than purely chain-like transmission; and \textbf{\emph{(nature)}} output-level preference can emerge earlier than representational consolidation, revealing a persistent gap between early bias and later commitment.

Our main contributions are threefold: \textbf{(1) the first causal, step-resolved evaluation view of latent CoT} that distinguishes when a solution becomes available from which steps remain causally necessary; \textbf{(2) an operator- and readout-conditioned influence analysis} that recovers dominant propagation routes while avoiding sparsity over-claims; and \textbf{(3) mode-conditional evidence that links latent-step budgets to practical design implications}: early “decision signals” need not imply early commitment, so improving latent reasoning likely requires shaping routing/commitment rather than simply adding more steps.

\section{Related Work}

\textbf{Latent and continuous chain-of-thought reasoning.} To avoid the cost and potential unfaithfulness of explicit CoT, recent methods move reasoning into continuous representations \citep{eit2025latentcot, zhu_survey_2025}. A first family performs depth-iterative latent reasoning, where the model recurrently updates a hidden state or continuous ``thought token" before decoding, as in Coconut \citep{hao_training_2024} and CODI \citep{shen_codi_2025}. Related approaches incorporate additional supervision or alignment to stabilize latent reasoning \citep{wei_sim-cot_2025, he_semcot_2025}, while hybrid methods mix latent and textual steps to balance efficiency and interpretability \citep{su_token_2025}. A second family scales test-time computation through recurrent-depth paradigms and parallel continuous updates \citep{geiping_scaling_2025, wu_parallel_2025, zhu2025scalinglatentreasoninglooped}. 
Most literature emphasizes accuracy and efficiency; our work focuses on internal causal organization within latent reasoning and how it relates to explicit CoT.

\textbf{Large reasoning models and reinforcement learning for reasoning.}
Recent work also improves LLM reasoning through reinforcement learning, selective supervision, uncertainty estimation, and long-horizon inference.
Representative directions include RL-based training and surveys for large reasoning models~\citep{zhang_survey_rl_lrm_2025}, fine-grained reward signals for test-time reinforcement learning~\citep{wang_beyond_majority_2025}, critical-token fine-tuning~\citep{ruan_critical_token_2025}, internal-belief-based uncertainty estimation~\citep{xiao_eagle_2025}, and infinite-horizon reasoning via reinforcement learning~\citep{yan_inftythink_plus_2026}.

\begin{figure*}[ht]
  \centering
  \includegraphics[width=\linewidth]{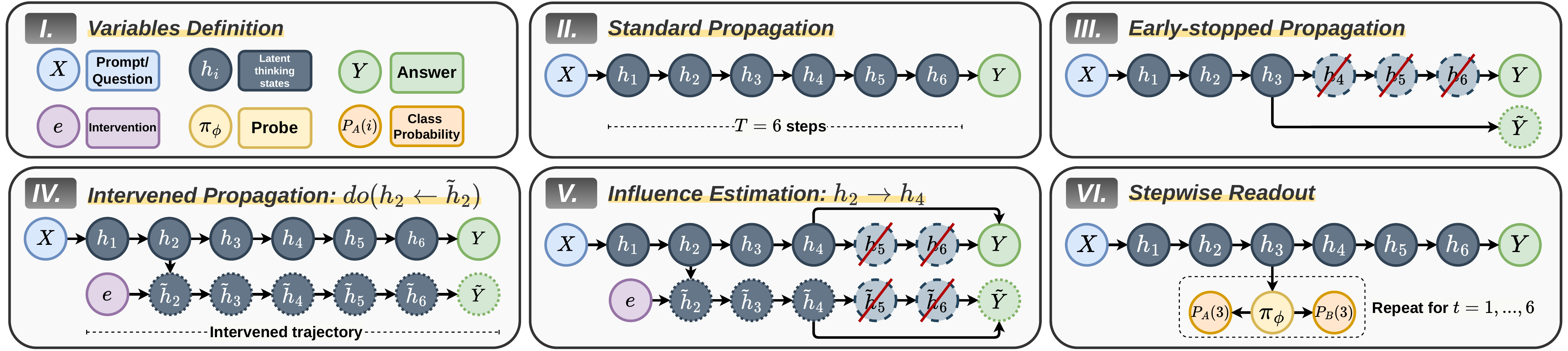}
  \caption{\textbf{Intervention-based protocol for latent CoT as a causal system.}
  \textbf{(I) Variables.} Input $X$ induces a latent trajectory $\{h_t\}_{t=1}^{T}$ and output $Y$; an intervention operator implements step-wise $\mathrm{do}(h_t \leftarrow \tilde h_t)$; a readout maps hidden states to answer support.
  \textbf{(II) Standard Propagation.} Unperturbed dynamics from $X$ through steps to $Y$.
  \textbf{(III) Early-stopped Propagation.} We truncate latent computation after step $k$ and decode from $h_k$ to test when correctness becomes decodable (RQ1).
  \textbf{(IV) Intervened Propagation.} We replace a step's state while keeping downstream computation intact, yielding an intervened outcome $\tilde Y$ (RQ1).
  \textbf{(V) Influence Estimation.} Combining a step-$t$ intervention with an early readout at step $s$ yields directed propagation strengths $W_{t,s}$ summarized as an empirical influence structure (RQ2).
  \textbf{(VI) Step-wise Readouts.} We read out the answer competition from $h_t$ (e.g., teacher forcing or a probe) to characterize superposition and commitment (RQ3).}
  \label{fig:method_overview}
  \vspace{-0.3cm}
\end{figure*}

\textbf{Chain-of-thought faithfulness and causal tests.}
A growing literature questions whether model-provided explanations faithfully reflect the computations that determine final predictions. Models can generate convincing yet unfaithful artifacts, including attention explanations, CoT rationales, and structured parses \citep{pruthi_learning_2020,turpin_language_2023,bai-etal-2025-constparse}. Prior work assesses faithfulness through context- or rationale-level interventions, such as deleting, shuffling, or editing rationales, and measures changes in downstream answers \citep{wang-etal-2023-causal,yang-etal-2023-causal,yu_causal_2025}. Complementary parameter-level interventions, such as unlearning reasoning steps from model weights, provide another causal probe of whether a step is internally represented and used \citep{tutek_measuring_2025}. 

\textbf{Causal and mechanistic analysis of internal representations.}
Causal perspectives view hidden activations as mediators that can be intervened on to test their functional role in model behavior \citep{pearl_causality_2022,feder_causal_2022,scholkopf_towards_2021}. Causal mediation analysis and mechanistic interventions have been used to distinguish behaviorally relevant internal components from merely correlated representations \citep{vig_investigating_2020,finlayson_causal_2021,stolfo_mechanistic_2023,serrano-smith-2019-attention,chrysostomou-aletras-2021-improving,elhage_toy_2022,nanda_progress_2023,meng_locating_2022,wang_interpretability_2023,conmy_towards_2023}. Recent work further shows that models can internalize reasoning strategies while concealing them from surface text under process supervision, highlighting the gap between latent computation and explicit explanation \citep{skaf_large_2025}. These toolkits motivate our step-level interventions on latent reasoning states.

\section{Evaluation Framework: Latent CoT as a Causal System}
\label{sec:framework}

\subsection{Scope and Causal Queries}
\label{sec:scope_queries}

We view latent CoT as a manipulable causal process evolving in continuous representation space.
Specifically, we treat intermediate latent reasoning steps as variables in a structural causal model (SCM) \citep{pearl_causality_2022,causal_survey}, which allows us to pose intervention-based causal queries and compute reproducible effect estimates under a fixed intervention protocol \citep{robins_identifiability_1992,imai_general_2010, vig_investigating_2020,mueller_quest_2026}.

Our analysis provides a unified intervention + readout protocol (Figure~\ref{fig:method_overview}) and uses it to answer three step-centric causal questions.
\textbf{(RQ1: necessity and sufficiency)} asks whether individual latent steps are behaviorally necessary, and whether the full latent budget is required to make the correct answer decodable.
\textbf{(RQ2: propagation and routing)} asks how a perturbation at an upstream step propagates to downstream computation, summarized as a step-to-step influence matrix and visualized as an empirical influence graph.
\textbf{(RQ3: superposition and commitment)} asks whether intermediate trajectories retain competing answer hypotheses and how output-level commitment relates to representational commitment across steps.

\subsection{Causal Variables, Minimal SCM, and Latent-step Interface}
\label{sec:scm_interface}

For an input problem $x$, we model latent reasoning as a sequence of continuous causal variables
$H_{1:T}=(H_1,\ldots,H_T)$ with $H_t\in\mathbb{R}^d$, where each $H_t$ corresponds to the model’s internal latent state
at reasoning step $t$. We denote the task-level output by $Y$ (e.g., the final predicted answer/label), and treat it as a random variable whose conditional distribution is induced by the model’s decoder given the latent trajectory.
A minimal SCM consistent with this computation is
\begin{align}
H_t &= f_t(H_{<t}, x, \epsilon_t; \theta),\quad t=1,\ldots,T, \\
Y   &= g(H_{1:T}, x, \epsilon_y; \theta),
\end{align}
where $f_t$ and $g$ are the model’s fixed transition and decoding mechanisms, and $\epsilon_t,\epsilon_y$ capture stochasticity.

In our experiments, we intervene on the intermediate states produced by a latent thinking rollout.
We denote realized latent states by lowercase $h_t$ and write
\begin{equation}
h_{1:T} \sim p_\theta(H_{1:T}\mid x),
\end{equation}
where $h_t\in\mathbb{R}^d$ instantiates the random variable $H_t$ at step $t$ under the model dynamics.
Latent-reasoning models expose these variables through a fixed-length sequence of hidden states
$h_{1:T}=(h_1,\ldots,h_T)$, where $h_t$ is the last-layer hidden representation associated with the $t$-th latent step
(e.g., a continuous ``thought token'' in \textsc{Coconut} or the designated reasoning position in \textsc{CODI}) and is used as the
step-$t$ reasoning input embedding.

\begin{definition}[Single-step latent $\mathrm{do}$-intervention]
\label{def:do_intervention}
Given input $x$ and a baseline rollout realizing $h_{1:T}$, the intervention
$\mathrm{do}(h_t \leftarrow \tilde h_t)$ replaces the step-$t$ latent state by $\tilde h_t$ and recomputes all downstream states using the same transition mechanism, yielding a counterfactual trajectory $\tilde h_{1:T}$.
Formally, let $\tilde h_{<t}=h_{<t}$ and $\tilde h_t$ be the overwritten state; for $t'>t$ we set
\begin{equation}
\tilde h_{t'} \;:=\; f_{t'}(\tilde h_{<t'}, x, \tilde\epsilon_{t'}; \theta),
\end{equation}
where $\tilde\epsilon_{t'}$ matches the baseline randomness when applicable.
The corresponding counterfactual output is obtained by the same readout mechanism,
\begin{equation}
\tilde y = g(\tilde h_{1:T}, x, \tilde\epsilon_y; \theta).
\end{equation}
\end{definition}

Unless otherwise stated, we use deterministic rollouts whenever possible; otherwise, we control randomness (e.g., fixed seeds) and
isolate propagation effects via teacher-forced readouts (Figure~\ref{fig:method_overview}(V)) to reduce sampling noise.

\subsection{Paradigms of Latent-reasoning Models}
\label{sec:overview_models}

We instantiate our framework on two latent-reasoning paradigms that differ in how they realize latent steps.

\textbf{Coconut} \citep{hao_training_2024} employs an explicit latent mode: it uses the final hidden state as a continuous reasoning token, feeding it back as the input for the next step, rather than decoding a discrete token.

\textbf{CODI} \citep{shen_codi_2025} compresses discrete CoT into continuous space via self-distillation: a continuous-CoT student is trained to both produce the correct answer and align its hidden states, at specific reasoning steps, with those of a discrete-CoT teacher, encouraging the latent trajectory to inherit stepwise structure.

\subsection{Models and Data}
\label{sec:models_data}

\paragraph{Models.}
We experiment with both \textsc{CODI} and \textsc{Coconut} on multiple backbones. We use official CODI checkpoints for GPT-2 \citep{radford_language_nodate} and Llama3-1B \citep{grattafiori_llama_2024}, and reproduce it on Qwen3-4B-Instruct \citep{yang_qwen3_2025}. Coconut models are reproduced across the same three backbones (GPT-2, Llama3-1B, Qwen3-4B-Instruct). Implementation details are in Appendix \ref{app:implementation}. Additional checks on Sim-CoT \citep{wei_sim-cot_2025} and larger/alternative \textsc{Coconut} backbones are reported in Appendix~\ref{app:additional_paradigm} and Appendix~\ref{app:additional_model_size}, respectively.

\paragraph{Datasets.}
We train and evaluate on CoT-augmented datasets that follow previous latent-reasoning methods.
For GSM8K, we train on GSM8K-Aug~\citep{deng_implicit_2023} and evaluate on the original GSM8K test set~\citep{cobbe_training_2021}.
For CommonsenseQA, we use the CoT-augmented training set released by \textsc{CODI}~\citep{shen_codi_2025} and evaluate on the original CommonsenseQA test set~\citep{talmor-etal-2019-commonsenseqa}.
Dataset details are provided in Appendix~\ref{app:data}.

\section{RQ1: Step-wise Necessity and Sufficiency}
\label{sec:rq1}

Latent reasoning replaces long textual rationales with a fixed-length sequence of intermediate hidden states, enabling step-wise
manipulation under the intervention-based protocol as introduced in Sec.~\ref{sec:framework} (Figure~\ref{fig:method_overview}).
RQ1 asks two causal questions aligned with this protocol: \textbf{(i) step-wise necessity,} i.e., whether the final decoded decision depends on a particular intermediate state, and \textbf{(ii) latent-budget sufficiency,} i.e., how many latent steps are required before the correct answer becomes decodable.
To address (i), we treat each realized latent state $h_t$ as an intervenable object and run paired rollouts that differ only by a single-step $\mathrm{do}$-intervention (Def.~\ref{def:do_intervention}) on $h_t$, quantifying its impact on the final decoded answer.
For (ii), we perform early-stop decoding by truncating latent computation after step $k$ and decoding directly from $h_k$ (Figure~\ref{fig:method_overview}(III)) to determine when correctness first becomes readable from the latent trajectory.

\begin{figure*}[ht]
  \centering
  \includegraphics[width=0.95\linewidth]{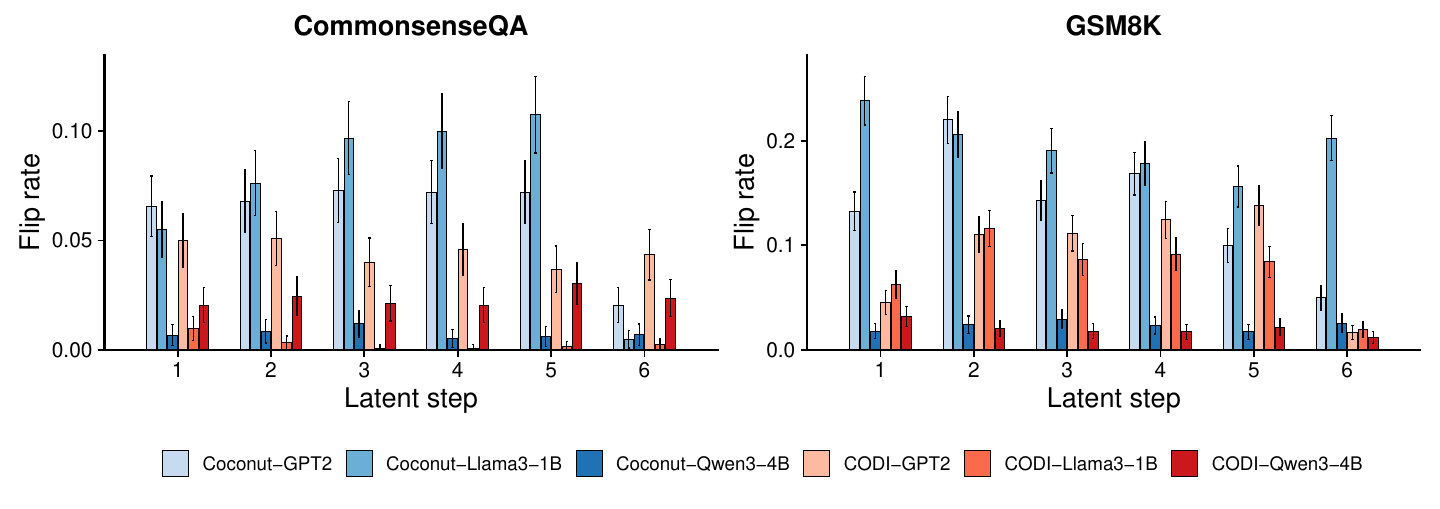}
  \vspace{-0.3cm}
  \caption{\textbf{Step-wise necessity measured by decision instability.}
  We intervene at a single latent step $t\in\{1,\dots,6\}$ by zeroing its state, $\mathrm{do}(h_t:=\mathbf{0})$, and then decode the final answer.
  We report the flip rate $\mathrm{Flip}(t)$, i.e., the fraction of examples whose decoded prediction changes relative to the baseline, on CommonsenseQA (left) and GSM8K (right).
  Error bars indicate estimation uncertainty.}
  \vspace{-0.3cm}
  \label{fig:rq1_flip_overview}
\end{figure*}

\subsection{Step-wise Interventions}
\label{sec:rq1_intervention}

\subsubsection{Experiment Setting}
\label{sec:rq1_intervention_setting}

We evaluate step-wise necessity using the step-wise $\mathrm{do}$-intervention in Def.~\ref{def:do_intervention}.
For each example, we run a baseline rollout and an intervened rollout that modifies exactly one latent state $h_t$ while keeping all
other components unchanged, including the input $x$, parameters $\theta$, and the downstream transition and readout mechanisms.
Following Appendix~\ref{app:intervention_operators}, we adopt the \textbf{zero intervention} for its simplicity and consistency across different models:
\begin{equation}
\mathrm{do}(h_t \leftarrow \tilde h_t),\quad \tilde h_t=\mathbf{0}.
\end{equation}
We then decode the final prediction $\tilde y^{(t)}$ and compute the \textbf{flip rate} $\mathrm{Flip}(t)$, defined as the fraction of
examples for which $\tilde y^{(t)}$ differs from the baseline output \citep{serrano-smith-2019-attention,chrysostomou-aletras-2021-improving}.
This metric directly measures decision-level dependence on the latent computation at step $t$ under a fixed intervention operator.

\subsubsection{Findings}
\label{sec:rq1_intervention_findings}
\textbf{Single-step ablation produces clear, step-specific flip patterns (not uniformly sensitive).}
Within the same model and dataset, $\mathrm{Flip}(t)$ changes noticeably with the intervened step $t$ (Figure~\ref{fig:rq1_flip_overview}), with several settings exhibiting mid-step peaks rather than a flat or monotone pattern. This indicates that different latent steps contribute differently to the final decision, and that single-step removal can selectively disrupt the decision more at some steps than others. 

\textbf{Arithmetic exhibits substantially higher decision volatility than commonsense.}
Flip rates on GSM8K are markedly higher than on CommonsenseQA for the same intervention protocol, with several backbones reaching flip rates around $0.1$--$0.2$ or higher on GSM8K, while CommonsenseQA remains mostly below $\sim 0.1$ (Figure~\ref{fig:rq1_flip_overview}). This gap is visible across both Coconut and CODI variants.

\textbf{Coconut shows larger flips than CODI under matched backbones, while stronger backbones suppress flips.}
Under the same backbone, Coconut variants generally yield higher flip rates than CODI, especially on GSM8K (Figure~\ref{fig:rq1_flip_overview}). Moreover, while stronger backbones substantially reduce flip rates across both paradigms, the flipping profile remains step-dependent—even when absolute rates are low.

\subsubsection{Implications}
These intervention results indicate that causal leverage is unevenly distributed across the latent trajectory. 
The effect of replacing a single latent state depends on where that state occurs and on the model--task setting in which it is used, rather than being an invariant property of the latent-step index. 
Thus, step-wise intervention identifies position-specific decision sensitivity within a fixed rollout, supporting a view of latent steps as differentiated components rather than uniformly interchangeable states.

\begin{figure}[t]
  \centering
  \includegraphics[width=\linewidth]{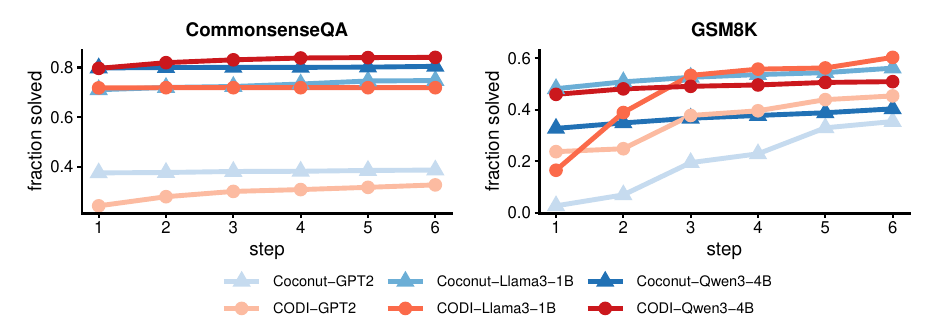}
  \caption{\textbf{Early-stop decoding reveals when correctness becomes decodable.}
  We report the cumulative solved fraction $S(k)=\mathbb{P}(k_i\le k)$ under early-stop decoding on CommonsenseQA (left) and GSM8K (right), where $k_i$ is the earliest step at which the correct answer becomes decodable (Def.~\ref{def:earliest_correct}).}
  \label{fig:rq1_earlystop_overview}
  \vspace{-0.3cm}
\end{figure}

\subsection{Early-stop Decoding}
\label{sec:rq1_earlystop}

\subsubsection{Experiment Setting}
\label{sec:rq1_earlystop_setting}

\begin{definition}[Earliest correct step under early-stop decoding]
\label{def:earliest_correct}
For each example $i$, let $\hat y_i^{(\le k)}$ be the decoded answer when we truncate latent computation after step $k$ and decode from $h_{i,k}$.
We define the earliest step at which the correct answer becomes decodable as
\begin{equation}
k_i = \min \{k \in \{1,\dots,T\}: \hat y_i^{(\le k)} = y_i^*\},
\end{equation}
and set $k_i=\infty$ if no early-stop decode is correct for $k\le T$.
\end{definition}

\begin{definition}[Cumulative solved fraction]
\label{def:earlystop_summaries}
Let $k_i$ be defined in Def.~\ref{def:earliest_correct}.
We define the cumulative solved fraction
\begin{equation}
S(k) = \frac{1}{N}\sum_{i=1}^{N}\mathbf{1}\{k_i \le k\}.
\end{equation}
\end{definition}

\subsubsection{Findings}
\label{sec:rq1_earlystop_findings}

\begin{figure*}[t]
  \centering
  \includegraphics[width=\textwidth]{figs/rq2/explicit_graph_grid.pdf}
  \caption{\textbf{Explicit CoT principal influence graphs (GSM8K; CoT-SFT baselines).}
  Nodes denote the first $T{=}6$ segmented CoT steps.
  Edge $t\!\to\!s$ indicates propagation strength $W_{t,s}$ from Eq.~\ref{eq:rq2_W} (teacher-forced KL shift on the gold answer under a single-step intervention at $t$ and readout at $s$).
  For readability, we show only top-1 outgoing edges after thresholding at $\alpha{=}0.1\cdot\max(W)$.}
  \label{fig:rq2_explicit_graphs}
\end{figure*}

\begin{figure*}[t]
  \centering
  \includegraphics[width=\textwidth]{figs/rq2/latent_graph_grid_gsm8k.pdf}
  \caption{\textbf{Latent principal influence graphs (GSM8K; \textsc{Coconut}/\textsc{CODI}).}
  Nodes are latent steps $t\in\{1,\dots,6\}$.
  Edge weights follow Eq.~\ref{eq:rq2_W} under single-step interventions, rendered with the same sparsification protocol as Figure~\ref{fig:rq2_explicit_graphs}.}
  \label{fig:rq2_latent_graphs}
    \vspace{-0.3cm}
\end{figure*}

\textbf{Early-stop curves differ across datasets.}
On CommonsenseQA, $S(k)$ typically rises rapidly within the first few steps and then saturates (Figure~\ref{fig:rq1_earlystop_overview}, left), suggesting most solvable instances become decodable within few steps.
In contrast, on GSM8K $S(k)$ often continues to increase toward later steps (Figure~\ref{fig:rq1_earlystop_overview}, right), with several settings showing gains up to $k{=}6$, indicating that additional latent computation can expand the set of instances for which the correct answer becomes decodable.

\textbf{Backbone strength shapes when correctness becomes decodable, while paradigm-level similarity is weak.}
Across both datasets, stronger backbones have higher $S(1)$ and earlier saturation of $S(k)$, whereas weaker backbones improve more gradually with $k$ (Figure~\ref{fig:rq1_earlystop_overview}).
Meanwhile, the step-wise profiles do not cluster by training paradigm: Coconut and CODI do not consistently exhibit similar curve shapes within each paradigm across backbones, suggesting that ``when correctness becomes decodable'' is not a stable paradigm-level signature in this experiment.

\subsubsection{Implications}
Early-stop decoding complements intervention analysis by measuring answer availability rather than step necessity.
The growth and saturation of $S(k)$ show that additional latent computation can expose correct-answer information for more instances, but the required budget depends on task difficulty and backbone capacity.
Thus, latent steps can be causally differentiated while still accumulating useful evidence across the trajectory.

\section{RQ2: Information Flow and Stepwise Influence Structure}
\label{sec:rq2}

RQ1 establishes that intervening on a single latent step can change the final decision, but it does not reveal where this perturbation propagates along the reasoning trajectory.
In RQ2, we compare step-to-step propagation of explicit CoT and latent CoT using an influence matrix $W$ (Eq.~\ref{eq:rq2_W}) and its principal influence graph rendering for readability.
We couple a single-step intervention at step $t$ with an early readout at a downstream step $s$, and measure how strongly the intervention changes the teacher-forced output distribution for the gold answer; see Figure~\ref{fig:method_overview} (V).
Unless otherwise stated, we report GSM8K in the main text; CommonsenseQA results are provided in Appendix~\ref{app:rq2_csqa}. Because the resulting graph is protocol-conditioned, we additionally test alternative intervention operators and an alternative readout protocol in Appendix~\ref{app:rq2_operator_robustness} and Appendix~\ref{app:rq2_logit_readout}.
\subsection{Experiment Setting}
\label{sec:rq2_setting}

\textbf{Nodes.}
For latent-reasoning models, each node corresponds to a latent step $t\in\{1,\dots,T\}$ with state $h_t$ (Sec.~\ref{sec:framework}).
For explicit CoT baselines (CoT-SFT), we segment the generated rationale into at most $T{=}6$ steps and represent step $t$ by the last-layer hidden state at the final token of that segment.\footnote{If an example has fewer than $T$ CoT steps, the remaining nodes are absent.}
This yields matched step-level trajectories for latent and explicit reasoning.

\textbf{Edges via intervention + early decoding.}
To probe directed propagation from step $t$ to a downstream step $s>t$, we run (i) a baseline rollout and (ii) an intervened rollout that modifies exactly one step state while keeping downstream computation unchanged.
We then \emph{decode at step $s$} using teacher forcing to obtain output distributions $p_{\text{base}}^{(s)}(\cdot)$ and $p_{\text{do}(t)}^{(s)}(\cdot)$, and define the example-level propagation strength as a position-averaged KL shift:
\begin{equation}
\mathrm{KL}^{(i)}_{t\to s}
=\frac{1}{|y_i^*|}\sum_{u=1}^{|y_i^*|}
\mathrm{KL}\!\left(
p_{\text{base}}^{(s)}(\cdot \mid y^*_{i,<u})
\ \Vert\
p_{\text{do}(t)}^{(s)}(\cdot \mid y^*_{i,<u})
\right).
\end{equation}
Aggregating over evaluation examples yields the influence matrix $W\in\mathbb{R}^{T\times T}$:
\begin{equation}
W_{t,s}=\mathbb{E}_i\big[\mathrm{KL}^{(i)}_{t\to s}\big],\quad t<s,
\label{eq:rq2_W}
\end{equation}
an operator-specific empirical influence structure under a fixed intervention and readout protocol, rather than a uniquely identified ``true'' causal graph.

\textbf{Principal influence graph rendering.}
For compact comparison across models, we visualize $W$ as a sparsified principal influence graph: we drop edges below $\alpha\cdot\max(W)$ with $\alpha{=}0.1$ and retain only the top-1 outgoing edge per node. 
Edge thickness scales with $W_{t,s}$, and dense heatmaps are provided in Appendix~\ref{app:rq2_full_mats}.

\textbf{Structural summaries.}
To quantify the heatmap-level patterns beyond visual inspection, we compute four normalized structure metrics on the dense $W$ (definitions in Appendix~\ref{app:rq2_metrics}): \textbf{locality} (mass near the diagonal), \textbf{span} (expected hop distance), \textbf{early-out} (influence originating from early steps), and \textbf{late-in} (influence terminating at late steps).
We normalize $W$ to remove scale effects across backbones before computing these metrics.

\begin{figure}[t]
  \centering
  \includegraphics[width=\linewidth]{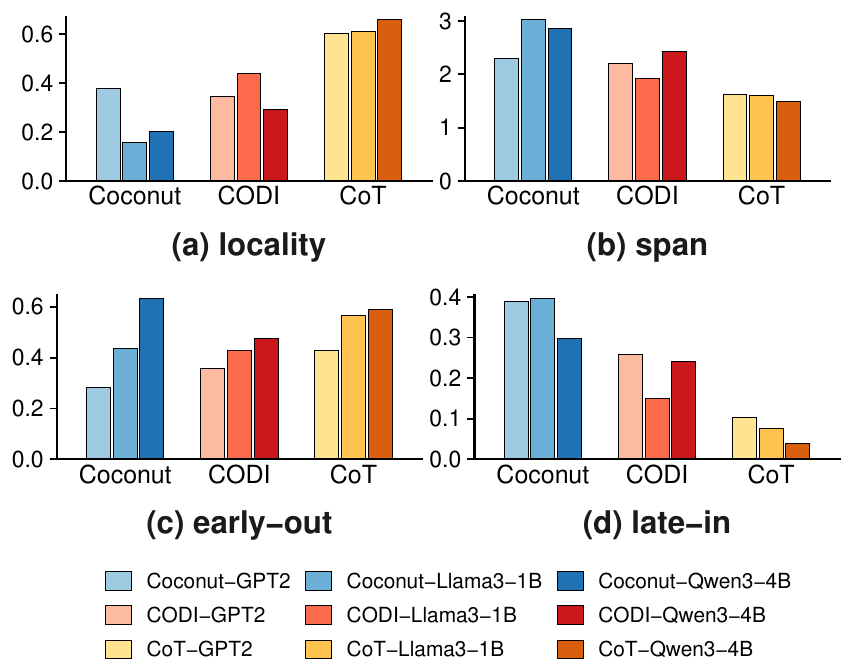}
  \caption{\textbf{Structural summaries of influence matrices (GSM8K).}
  We report locality, span, early-out, and late-in computed on dense normalized $W$ (Appendix~\ref{app:rq2_metrics}).}
  \label{fig:rq2_structure_metrics}
  \vspace{-0.3cm}
\end{figure}

\subsection{Findings}
\label{sec:rq2_findings}

\textbf{Explicit CoT influence graphs remain near-chain across backbones.}
CoT-SFT exhibits a consistently sequential topology: the dominant edges follow adjacent transitions, with only minor deviations across backbones (Figure~\ref{fig:rq2_explicit_graphs}).
This stability is also reflected in the structure summaries on GSM8K (Figure~\ref{fig:rq2_structure_metrics}): CoT-SFT has uniformly high locality (all $\ge 0.6$) with low span, matching the intuition that textual steps induce predominantly local dependencies.

\textbf{Latent influence graphs are dominated by skip connections, revealing non-chain propagation.}
Latent graphs contain substantially more skip connections than explicit CoT (Figure~\ref{fig:rq2_latent_graphs}), indicating that influence often bypasses intermediate steps rather than accumulating strictly along a local chain.
This is also captured by the structure summaries on GSM8K (Figure~\ref{fig:rq2_structure_metrics}): latent models are markedly less local and have larger spans than CoT-SFT, and they place substantially more normalized influence into late-step targets (late-in).
Within this skip-dominant regime, \textsc{Coconut} tends to exhibit more pronounced early$\to$late routing (often connecting early steps directly to the final step), while \textsc{CODI} departs from a strict chain but is generally less dominated by early$\to$final shortcuts and shows greater variation across backbones (Figure~\ref{fig:rq2_latent_graphs}). Despite this difference between the two paradigms, the skip connections, together with the non-uniform sensitivity observed in Figure~\ref{fig:rq1_flip_overview}, indicate functional differentiation across nodes in the latent trajectory. 

\subsection{Implications}
These influence patterns suggest that latent reasoning does not simply preserve the causal flow of the explicit CoT trajectories from which it is derived. Although COCONUT and CODI both obtain latent computation by replacing or compressing explicit reasoning steps into continuous states, their step-to-step influence structures depart from the near-chain topology observed in explicit CoT. Under the same intervention and readout protocol, explicit CoT remains close to local sequential propagation, whereas latent CoT exposes longer-range routes that often terminate at later readouts.  This implies that latentization changes not only the surface format of reasoning, but also its internal routing structure: computational adjacency in latent space need not mirror textual adjacency in natural-language rationales.

\section{RQ3: Superposition and Commitment in Latent Dynamics}
\label{sec:rq3}

RQ1 demonstrates that removing a single latent step can alter the final decision, while RQ2 reveals step-to-step influence is non-local: early computation can directly affect multiple later steps, bypassing adjacent ones. 
A remaining ambiguity is whether such non-locality reflects (i) \emph{early commitment} to one answer mode, which is then propagated, or (ii) \emph{sustained competition} among multiple hypotheses within the latent trajectory.
Prior work offers contrasting views: some argue continuous ``soft thinking'' remains effectively greedy and single-threaded \citep{wu_llms_2025}, while others suggest latent reasoning may retain competing hypotheses in shared subspaces \citep{zhu_reasoning_2025}.

RQ3 therefore asks a trajectory-level question:
\textbf{when stochastic rollouts of the same prompt lead to different final answers, do intermediate steps exhibit superposition, and how does this evolve across steps?}
We instantiate this analysis on \textbf{StrategyQA}~\citep{geva-etal-2021-aristotle}, whose binary label space (\texttt{Yes}/\texttt{No}) clearly defines two modes and allows direct probability-based readout at each step. For GSM8K, whose open-ended numeric answers make two-mode filtering substantially sparser, we provide a supporting analysis in Appendix~\ref{app:rq3_gsm8k}.

\subsection{Experiment Setting}
\label{sec:rq3_setting}

\textbf{Two-mode prompts from stochastic rollouts.}
For each prompt $x$, we enable stochastic decoding and sample $K$ rollouts.
Each rollout produces a latent trajectory $\{h^{(k)}_t\}_{t=1}^{T}$ and a final answer $\hat y^{(k)}\in\{\texttt{Yes},\texttt{No}\}$.
We retain prompts whose rollouts contain both answers, and partition rollouts into $\mathcal{C}_Y$ and $\mathcal{C}_N$ accordingly.
Filtering thresholds and the filtering strategy are reported in Appendix~\ref{app:rq3_stats}.

\textbf{Intermediate-step readouts (teacher-forced vs.\ probe).}
At each latent step $t$, we quantify the model's relative support for \texttt{Yes} vs.\ \texttt{No} using two readouts: (i) \textbf{teacher-forced} scoring under a fixed answer template, and (ii) a lightweight \textbf{probe} trained on frozen latent states to predict the rollout mode. Both yield a step-wise binary distribution over the two answer modes, denoted by $p_Y(t)$ and $p_N(t)$. Templates and probing details are provided in Appendix~\ref{app:rq3_readout}.

\textbf{Superposition score.}
Given $p_Y(t),p_N(t)$, we define the step-wise superposition score
\begin{equation}
\mathrm{SS}(t)=\min\big(p_Y(t),p_N(t)\big),
\end{equation}
which is high when both answers retain non-trivial support and low when one answer dominates.

\begin{figure}[t]
  \centering
  \includegraphics[width=\linewidth]{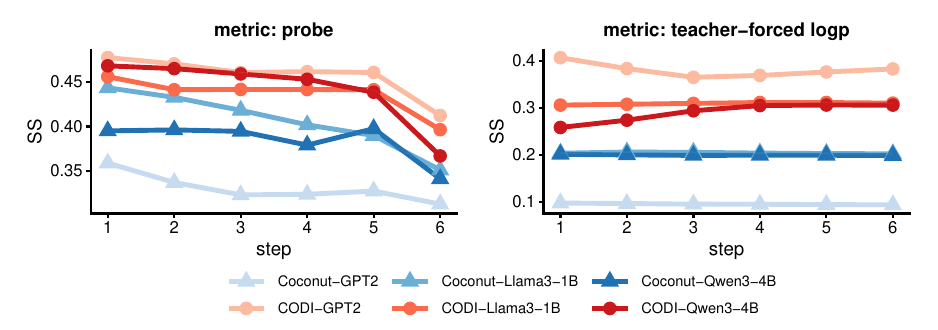}
  \caption{\textbf{RQ3: Step-wise superposition on StrategyQA.}
  Left: probe readout; right: teacher-forced log-probability readout.
  The superposition score $\mathrm{SS}$ measures how much support is simultaneously retained for \texttt{Yes} and \texttt{No} at each latent step.}
  \label{fig:rq3_superposition_main}
  \vspace{-0.3cm}
\end{figure}

\subsection{Findings}
\label{sec:rq3_findings}

\textbf{Teacher-forced readout suggests early output-level commitment.}
In Figure~\ref{fig:rq3_superposition_main} (right), $\mathrm{SS}(t)$ is uniformly low and varies only modestly across steps, indicating that the model’s answer distribution becomes skewed toward either \texttt{Yes} or \texttt{No} early in the latent budget.
A minimal implication is that the non-local edges in RQ2 can be compatible with early mode selection followed by propagation, without requiring prolonged output-level ambiguity (cf.\ the non-local principal influence graphs in Figure~\ref{fig:rq2_latent_graphs}).

\textbf{Probe readout reveals more sustained competition and a late collapse.}
In Figure~\ref{fig:rq3_superposition_main} (left), the probe panel shows substantially higher $\mathrm{SS}(t)$ throughout the trajectory, with a clear drop at the final step, implying that intermediate states can retain decodable support for the alternative mode even when teacher forcing appears committed.
This readout gap highlights readout dependence: intermediate representations may remain ``multi-mode available'' beyond what is expressed by the default answer distribution.

\textbf{Paradigm-level separation aligns with RQ2's structure differences.}
Under the probe readout, \textsc{CODI} variants maintain higher superposition scores than \textsc{Coconut} variants across latent steps (Figure~\ref{fig:rq3_superposition_main}, left).
This separation is consistent with RQ2: \textsc{Coconut} more often exhibits dominant early-to-late principal edges, whereas \textsc{CODI} shows less extreme long-range routing (Figures~\ref{fig:rq2_latent_graphs} and~\ref{fig:rq2_structure_metrics}).

\subsection{Implications}
These readouts distinguish output-level commitment from representational commitment. Teacher-forced scoring indicates that the default answer distribution can become biased toward one mode early, while probe readout shows that alternative modes can remain accessible in the intermediate states. Thus, commitment in latent reasoning is not a single collapse event: output preference may emerge before the latent trajectory has fully eliminated competing representational modes.

\section{Discussion}
\label{sec:discussion}

Our experiments connect step-wise necessity and early decodability (RQ1), directed propagation (RQ2), and trajectory-level mode dynamics (RQ3). Across \textsc{Coconut} and \textsc{CODI}, the results suggest that a fixed latent budget behaves less like homogeneous extra depth than a structured interface: steps have unequal causal leverage, influence can route non-locally, and output-level commitment need not coincide with representational commitment.

\textbf{Latent steps are causally functional, with heterogeneous leverage.}
RQ1 shows that latent computation is not merely incidental to decoding: single-step interventions can alter the final decision, but this causal dependence is unevenly distributed across the latent budget.
This pattern suggests an implicit \emph{division of labor}, where some steps act as high-leverage intervention sites while others contribute more conditionally, depending on the input or reasoning mode.
Rather than repeatedly refining a homogeneous state, latent reasoning appears to introduce step-specific updates whose downstream effects are later amplified, transformed, or gated.
The different leverage profiles of \textsc{Coconut} and \textsc{CODI} further suggest that the training paradigm shapes where decision-relevant dependence concentrates.

\textbf{Principal influence structure reveals functional routes of latent computation.}
Step-wise necessity profiles alone do not identify \emph{how} a perturbation travels through the remaining latent trajectory.
The principal influence graphs from RQ2 add this missing structure by highlighting dominant directed routes through which an intervention at step $t$ manifests at later readouts.
A recurring implication is that latent computation can be organized around a small number of effective long-range paths, where early steps shape later representations without requiring strong adjacent mediation at every intermediate step.
This structural view helps reconcile cases where a step exhibits modest direct necessity but still participates in a high-influence route: its role may be to shape downstream states that only become consequential after later consolidation.

\textbf{Superposition and commitment: shared computation versus collapse.}
Early-stop decoding measures when the correct answer first becomes readable, whereas intervention sensitivity and RQ3 readouts reveal whether later computation remains behaviorally and representationally relevant.
Together, they separate \emph{availability} from \emph{stability}: an answer may be decodable before the latent trajectory consolidates the decision or eliminates competing modes.
Through RQ2, this suggests that superposition can reflect \emph{shared prefix computation}, where competing solutions reuse intermediate processing before divergence becomes externally visible.
Probe-based and teacher-forced readouts further show that commitment is not a single event: output distributions can collapse before intermediate representations cease to carry alternatives.
This aligns with RQ1's late-step stabilization effects and with evidence that behavioral or preference directions can be isolated and steered in representation space via lightweight inference-time interventions \citep{rimsky-etal-2024-steering, 10.5555/3666122.3667919, zou2025representationengineeringtopdownapproach, zhang2026evaluatingsteeringmodalitypreferences}.
Overall, RQ1--RQ3 suggest a division of labor across latent steps, where shared intermediate computation and paradigm-dependent stabilization mechanisms support robust final-mode dominance.

\textbf{Implications for latent-reasoning design.}
Although our contribution is diagnostic rather than algorithmic, the observed step-level structure suggests concrete design directions.
Latent budgets should not be treated as homogeneous extra depth: computation, supervision, and regularization should be allocated selectively according to the functional role of each step.
Since latent states lack natural supervision targets, future objectives could move beyond uniform CoT imitation and instead impose functional constraints on high-leverage steps or high-influence routes, such as readout consistency, perturbation robustness, downstream-state stability, or teacher-derived targets at causally important positions.
The gap between early answer preference and later representational commitment further argues against forcing every step to predict the final answer: early states should preserve uncertainty or multiple candidate modes, while later states consolidate them into a perturbation-stable decision.
Accordingly, stopping rules should consider not only when an answer first becomes decodable, but also when the supporting latent representation becomes stable.

\section{Conclusion}
We framed latent chain-of-thought as a step-indexed causal system and evaluated it across \textsc{Coconut} and \textsc{CODI} with interventions, influence-structure estimation, and trajectory-level readouts. Beyond characterizing these models, our results offer design-relevant insights: latent-step budgets should be treated as an allocatable interface rather than homogeneous ``extra depth,'' since causal leverage concentrates unevenly and propagates along a few dominant long-range routes; and training/decoding should account for a gap between early output bias and late representational commitment, where alternatives remain latent-available even after the output distribution tilts. Together, this suggests that future latent-reasoning models can be improved by explicitly shaping where information is written and how it is consolidated across steps (e.g., encouraging more stable bottlenecks or controllable routing), rather than only scaling the number of latent steps.

\section*{Limitations}
Our conclusions are tied to specific methodological choices, including the step-level causal interface, hidden-state overwrite interventions, and teacher-forced readout.
Furthermore, this study is limited to single-step edits, a fixed latent budget ($T{=}6$), and a controlled set of paradigms, backbones, and CoT-supervised benchmarks. Although we provide supplementary checks on \textsc{Sim-CoT} and additional \textsc{Coconut} backbones in Appendix~\ref{app:additional_paradigm} and Appendix~\ref{app:additional_model_size}, evaluation across more paradigms, longer horizons, adaptive latent budgets, and varied intervention types is still needed to assess the generalizability of the observed patterns.

\newpage
\section*{Acknowledgments}
This work was supported in part by the Science Fund for Creative Research Groups of the National Natural Science Foundation of China (62521006), in part by the National Natural Science Foundation of China (62406091, 62276077, U23B2055, 62350710797),  in part by Guangdong S\&T Program (2024B0101050003), in part by the Guangdong Basic and Applied Basic Research Foundation (2026A1515011718, 2024A1515011205), in part by the Shenzhen Science and Technology Program (KQTD20240729102154066), in part by the Fundamental Research Funds for the Central Universities (GW2025-19) and in part by the State Key Laboratory of Complex \& Critical Software Environment (SKLCCSE-2025ZX-26).

\section*{Impact Statement}
This work studies how latent-reasoning language models can be analyzed using intervention-based measurements. The primary impact is methodological: providing tools to better diagnose, compare, and potentially improve the reliability and controllability of latent computation. While improved latent reasoning could enable more capable automated decision-making systems, the same advances may also lower the cost of generating persuasive or overconfident outputs. We therefore emphasize responsible use: evaluations should be paired with standard safety testing, and interpretations of causal measurements should avoid overclaiming model understanding beyond what the protocol supports.


\bibliography{example_paper}

@article{cobbe_training_2021,
	author = {Cobbe, Karl and Kosaraju, Vineet and Bavarian, Mohammad and Chen, Mark and Jun, Heewoo and Kaiser, Lukasz and Plappert, Matthias and Tworek, Jerry and Hilton, Jacob and Nakano, Reiichiro and Hesse, Christopher and Schulman, John},
	title = {Training {Verifiers} to {Solve} {Math} {Word} {Problems}},
	journal = {arXiv preprint arXiv:2110.14168},
	year = {2021},
	doi = {10.48550/arXiv.2110.14168},
	url = {https://doi.org/10.48550/arXiv.2110.14168},
}

@inproceedings{wei_chain--thought_2022,
	author = {Jason Wei and Xuezhi Wang and Dale Schuurmans and Maarten Bosma and Brian Ichter and Fei Xia and Ed H. Chi and Quoc V. Le and Denny Zhou},
	title = {{Chain-of-Thought} {Prompting} {Elicits} {Reasoning} in {Large} {Language} {Models}},
	booktitle = {Advances in {Neural} {Information} {Processing} {Systems}},
	year = {2022},
	month = dec,
	address = {New Orleans, LA, USA},
	volume = {35},
	pages = {24824--24837},
	publisher = {Curran Associates, Inc.},
    url = {https://proceedings.neurips.cc/paper_files/paper/2022/file/9d5609613524ecf4f15af0f7b31abca4-Paper-Conference.pdf},
}

@inproceedings{pruthi_learning_2020,
	author = {Pruthi, Danish and Gupta, Mansi and Dhingra, Bhuwan and Neubig, Graham and Lipton, Zachary C.},
	title = {Learning to Deceive with Attention-Based Explanations},
	booktitle = {Proceedings of the 58th Annual Meeting of the Association for Computational Linguistics},
	year = {2020},
	month = jul,
	address = {Online},
	pages = {4782--4793},
	publisher = {Association for Computational Linguistics},
	doi = {10.18653/v1/2020.acl-main.432},
	url = {https://aclanthology.org/2020.acl-main.432/},
}

@inproceedings{turpin_language_2023,
	author = {Turpin, Miles and Michael, Julian and Perez, Ethan and Bowman, Samuel R.},
	title = {{Language} {Models} Don\textquotesingle t {Always} {Say} {What} {They} {Think}: {Unfaithful} {Explanations} in {Chain-of-Thought} {Prompting}},
	booktitle = {Advances in {Neural} {Information} {Processing} {Systems}},
	year = {2023},
	month = dec,
	address = {New Orleans, LA, USA},
	volume = {36},
	pages = {74952--74965},
	publisher = {Curran Associates, Inc.},
	url = {https://proceedings.neurips.cc/paper_files/paper/2023/file/ed3fea9033a80fea1376299fa7863f4a-Paper-Conference.pdf},
}

@inproceedings{nanda_progress_2023,
	author = {Neel Nanda and Lawrence Chan and Tom Lieberum and Jess Smith and Jacob Steinhardt},
	title = {Progress {Measures} for {Grokking} via {Mechanistic} {Interpretability}},
	booktitle = {The {Eleventh} {International} {Conference} on {Learning} {Representations}},
	year = {2023},
	month = may,
	address = {Kigali, Rwanda},
	url = {https://openreview.net/forum?id=9XFSbDPmdW},
}

@article{zhu_survey_2025,
	author = {Zhu, Rui-Jie and Peng, Tianhao and Cheng, Tianhao and Qu, Xingwei and Huang, Jinfa and Zhu, Dawei and Wang, Hao and Xue, Kaiwen and Zhang, Xuanliang and Shan, Yong and Cai, Tianle and Kergan, Taylor and Kembay, Assel and Smith, Andrew and Lin, Chenghua and Nguyen, Binh and Pan, Yuqi and Chou, Yuhong and Cai, Zefan and Wu, Zhenhe and Zhao, Yongchi and Liu, Tianyu and Yang, Jian and Zhou, Wangchunshu and Zheng, Chujie and Li, Chongxuan and Zhou, Yuyin and Li, Zhoujun and Zhang, Zhaoxiang and Liu, Jiaheng and Zhang, Ge and Huang, Wenhao and Eshraghian, Jason},
	title = {A {Survey} on {Latent} {Reasoning}},
	journal = {arXiv preprint arXiv:2507.06203},
	year = {2025},
	doi = {10.48550/arXiv.2507.06203},
	url = {https://doi.org/10.48550/arXiv.2507.06203},
}

@inproceedings{hao_training_2024,
	author = {Hao, Shibo and Sukhbaatar, Sainbayar and Su, DiJia and Li, Xian and Hu, Zhiting and Weston, Jason E. and Tian, Yuandong},
	title = {Training {Large} {Language} {Models} to {Reason} in a {Continuous} {Latent} {Space}},
	booktitle = {Second {Conference} on {Language} {Modeling}},
	year = {2025},
	month = oct,
	address = {Montreal, Canada},
	url = {https://openreview.net/forum?id=Itxz7S4Ip3},
}

@inproceedings{geiping_scaling_2025,
	author = {Geiping, Jonas and McLeish, Sean Michael and Jain, Neel and Kirchenbauer, John and Singh, Siddharth and Bartoldson, Brian R. and Kailkhura, Bhavya and Bhatele, Abhinav and Goldstein, Tom},
	title = {Scaling up {Test-Time} {Compute} with {Latent} {Reasoning}: {A} {Recurrent} {Depth} {Approach}},
	booktitle = {Advances in {Neural} {Information} {Processing} {Systems}},
	year = {2025},
	month = dec,
	address = {San Diego, CA, USA and Mexico City, Mexico},
	volume = {38},
    pages = {41340--41391},
	publisher = {Curran Associates, Inc.},
    url = {https://proceedings.neurips.cc/paper_files/paper/2025/file/3b01972cf31e6fa0fe29e4b8b5c2a0a1-Paper-Conference.pdf},
}

@inproceedings{xu_softcot_2025,
	author = {Xu, Yige and Guo, Xu and Zeng, Zhiwei and Miao, Chunyan},
	title = {{SoftCoT}: {Soft} {Chain}-of-{Thought} for {Efficient} {Reasoning} with {LLMs}},
	booktitle = {Proceedings of the 63rd {Annual} {Meeting} of the {Association} for {Computational} {Linguistics} ({Volume} 1: {Long} {Papers})},
	year = {2025},
	month = jul,
	address = {Vienna, Austria},
	pages = {23336--23351},
	publisher = {Association for Computational Linguistics},
	doi = {10.18653/v1/2025.acl-long.1137},
	url = {https://aclanthology.org/2025.acl-long.1137/},
}

@inproceedings{shen_codi_2025,
	author = {Shen, Zhenyi and Yan, Hanqi and Zhang, Linhai and Hu, Zhanghao and Du, Yali and He, Yulan},
	title = {{CODI}: {Compressing} {Chain}-of-{Thought} into {Continuous} {Space} via {Self}-{Distillation}},
	booktitle = {Proceedings of the 2025 {Conference} on {Empirical} {Methods} in {Natural} {Language} {Processing}},
	year = {2025},
	month = nov,
	address = {Suzhou, China},
	pages = {677--693},
	publisher = {Association for Computational Linguistics},
	doi = {10.18653/v1/2025.emnlp-main.36},
	url = {https://aclanthology.org/2025.emnlp-main.36/},
}

@inproceedings{tutek_measuring_2025,
	author = {Tutek, Martin and Hashemi Chaleshtori, Fateme and Marasovic, Ana and Belinkov, Yonatan},
	title = {Measuring {Chain} of {Thought} {Faithfulness} by {Unlearning} {Reasoning} {Steps}},
	booktitle = {Proceedings of the 2025 {Conference} on {Empirical} {Methods} in {Natural} {Language} {Processing}},
	year = {2025},
	month = nov,
	address = {Suzhou, China},
	pages = {9935--9960},
	publisher = {Association for Computational Linguistics},
	doi = {10.18653/v1/2025.emnlp-main.504},
	url = {https://aclanthology.org/2025.emnlp-main.504/},
}

@inproceedings{wei_sim-cot_2025,
	author = {Wei, Xilin and Liu, Xiaoran and Zang, Yuhang and Dong, Xiaoyi and Cao, Yuhang and Wang, Jiaqi and Qiu, Xipeng and Lin, Dahua},
	title = {{SIM}-{C}o{T}: {Supervised} {Implicit} {Chain-of-Thought}},
	booktitle = {The {Fourteenth} {International} {Conference} on {Learning} {Representations}},
	year = {2026},
	month = apr,
	address = {Rio de Janeiro, Brazil},
	url = {https://openreview.net/forum?id=6YRJ4jmVQl},
}

@inproceedings{wu_parallel_2025,
	author = {Wu, Haoyi and Teng, Zhihao and Tu, Kewei},
	title = {Parallel {Continuous} {Chain}-of-{Thought} with {Jacobi} {Iteration}},
	booktitle = {Proceedings of the 2025 {Conference} on {Empirical} {Methods} in {Natural} {Language} {Processing}},
	year = {2025},
	month = nov,
	address = {Suzhou, China},
	pages = {914--926},
	publisher = {Association for Computational Linguistics},
	doi = {10.18653/v1/2025.emnlp-main.47},
	url = {https://aclanthology.org/2025.emnlp-main.47/},
}

@inproceedings{su_token_2025,
	author = {DiJia Su and Hanlin Zhu and Yingchen Xu and Jiantao Jiao and Yuandong Tian and Qinqing Zheng},
	title = {{Token Assorted}: {Mixing} {Latent} and {Text} {Tokens} for {Improved} {Language} {Model} {Reasoning}},
	booktitle = {Proceedings of the 42nd {International} {Conference} on {Machine} {Learning}},
	year = {2025},
	month = jul,
	address = {Vancouver, BC, Canada},
	volume = {267},
	pages = {57144--57163},
	series = {Proceedings of {Machine} {Learning} {Research}},
	publisher = {{PMLR}},
	url = {https://proceedings.mlr.press/v267/su25g.html},
}

@inproceedings{wu_llms_2025,
	author = {Wu, Junhong and Lu, Jinliang and Ren, Zixuan and Hu, Gangqiang and Wu, Zhi and Dai, Dai and Wu, Hua},
	title = {{{LLM}s} are {Single-threaded} {Reasoners}: {Demystifying} the {Working} {Mechanism} of {Soft} {Thinking}},
	booktitle = {The {Fourteenth} {International} {Conference} on {Learning} {Representations}},
	year = {2026},
	month = apr,
	address = {Rio de Janeiro, Brazil},
	url = {https://openreview.net/forum?id=ASLuOoP78o},
}

@inproceedings{zhu_reasoning_2025,
	author = {Zhu, Hanlin and Hao, Shibo and Hu, Zhiting and Jiao, Jiantao and Russell, Stuart J and Tian, Yuandong},
	title = {{Reasoning} by {Superposition}: {A} {Theoretical} {Perspective} on {Chain} of {Continuous} {Thought}},
	booktitle = {Advances in {Neural} {Information} {Processing} {Systems}},
	year = {2025},
	month = dec,
	address = {San Diego, CA, USA and Mexico City, Mexico},
	volume = {38},
	pages = {79931--79963},
	publisher = {Curran Associates, Inc.},
	url = {https://proceedings.neurips.cc/paper_files/paper/2025/file/72c363c2a573ca2128bd176d3317696b-Paper-Conference.pdf},
}

@inproceedings{skaf_large_2025,
	author = {McCarthy, Robert and SKAF, Joey and Ibanez-Lissen, Luis and Georgiev, Vasil and Watts, Connor and Whittingham, Hannes and Gonzalez-Manzano, Lorena and Tice, Cameron and Young, Edward James and Radmard, Puria and Lindner, David},
	title = {Large {Language} {Models} Can {Learn} and {Generalize} {Steganographic} {Chain-of-Thought} under {Process} {Supervision}},
	booktitle = {Advances in {Neural} {Information} {Processing} {Systems}},
	year = {2025},
	month = dec,
	address = {San Diego, CA, USA and Mexico City, Mexico},
	volume = {38},
    pages = {27597--27632},
	publisher = {Curran Associates, Inc.},
	url = {https://proceedings.neurips.cc/paper_files/paper/2025/file/28131b22fafebba500eb7bb02e3d5b59-Paper-Conference.pdf},
}

@inproceedings{he_semcot_2025,
	author = {He, Yinhan and Zheng, Wendy and Zhu, Yaochen and Zheng, Zaiyi and Su, Lin and Vasudevan, Sriram and Guo, Qi and Hong, Liangjie and Li, Jundong},
	title = {{SemCoT}: {Accelerating} {Chain-of-Thought} {Reasoning} through {Semantically-Aligned} {Implicit} {Tokens}},
	booktitle = {Advances in {Neural} {Information} {Processing} {Systems}},
	year = {2025},
	month = dec,
	address = {San Diego, CA, USA and Mexico City, Mexico},
	volume = {38},
    pages = {43455--43485},
	publisher = {Curran Associates, Inc.},
	url = {https://proceedings.neurips.cc/paper_files/paper/2025/file/3ddbd473456a57e3cafb1ee51ddf8ff6-Paper-Conference.pdf},

}

@inproceedings{gozeten_continuous_2025,
	author = {Gozeten, Halil Alperen and Ildiz, Muhammed Emrullah and Zhang, Xuechen and Harutyunyan, Hrayr and Rawat, Ankit Singh and Oymak, Samet},
	title = {Continuous {Chain} of {Thought} {Enables} {Parallel} {Exploration} and {Reasoning}},
	booktitle = {The {Fourteenth} {International} {Conference} on {Learning} {Representations}},
	year = {2026},
	month = apr,
	address = {Rio de Janeiro, Brazil},
	url = {https://openreview.net/forum?id=sTPKDKn5ig},
}

@article{scholkopf_towards_2021,
	author = {Sch{\"o}lkopf, Bernhard and Locatello, Francesco and Bauer, Stefan and Ke, Nan Rosemary and Kalchbrenner, Nal and Goyal, Anirudh and Bengio, Yoshua},
	title = {Toward {Causal} {Representation} {Learning}},
	journal = {Proceedings of the {IEEE}},
	year = {2021},
	volume = {109},
	number = {5},
	pages = {612--634},
	publisher = {Institute of Electrical and Electronics Engineers},
	doi = {10.1109/JPROC.2021.3058954},
	url = {https://doi.org/10.1109/JPROC.2021.3058954},
}

@article{feder_causal_2022,
	author = {Feder, Amir and Keith, Katherine A. and Manzoor, Emaad and Pryzant, Reid and Sridhar, Dhanya and Wood-Doughty, Zach and Eisenstein, Jacob and Grimmer, Justin and Reichart, Roi and Roberts, Margaret E. and Stewart, Brandon M. and Veitch, Victor and Yang, Diyi},
	title = {Causal {Inference} in {Natural} {Language} {Processing}: {Estimation}, {Prediction}, {Interpretation} and {Beyond}},
	journal = {Transactions of the Association for Computational Linguistics},
	year = {2022},
	volume = {10},
	pages = {1138--1158},
	doi = {10.1162/tacl_a_00511},
	url = {https://aclanthology.org/2022.tacl-1.66/},
}

@article{elhage_toy_2022,
	author = {Elhage, Nelson and Hume, Tristan and Olsson, Catherine and Schiefer, Nicholas and Henighan, Tom and Kravec, Shauna and Hatfield-Dodds, Zac and Lasenby, Robert and Drain, Dawn and Chen, Carol and Grosse, Roger and McCandlish, Sam and Kaplan, Jared and Amodei, Dario and Wattenberg, Martin and Olah, Christopher},
	title = {Toy {Models} of {Superposition}},
	journal = {arXiv preprint arXiv:2209.10652},
	year = {2022},
	doi = {10.48550/arXiv.2209.10652},
	url = {https://doi.org/10.48550/arXiv.2209.10652},
}

@inproceedings{zhang_soft_2025,
	author = {Zhang, Zhen and He, Xuehai and Yan, Weixiang and Shen, Ao and Zhao, Chenyang and Wang, Xin},
	title = {{Soft} {Thinking}: {Unlocking} the {Reasoning} {Potential} of {{LLM}s} in {Continuous} {Concept} {Space}},
	booktitle = {Advances in {Neural} {Information} {Processing} {Systems}},
	year = {2025},
	month = dec,
	address = {San Diego, CA, USA and Mexico City, Mexico},
	volume = {38},
	pages = {168990--169012},
	publisher = {Curran Associates, Inc.},
	url = {https://proceedings.neurips.cc/paper_files/paper/2025/file/f7396d1c54d51416958d63e285377103-Paper-Conference.pdf},
}

@book{pearl_causality_2022,
	address = {Cambridge New York, NY Port Melbourne New Delhi Singapore},
	edition = {Second edition, reprinted with corrections},
	title = {Causality: models, reasoning, and inference},
	isbn = {978-0-521-89560-6},
	language = {eng},
	publisher = {Cambridge University Press},
	author = {Pearl, Judea},
	year = {2000},
}

@article{yang_qwen3_2025,
	author = {Yang, An and Li, Anfeng and Yang, Baosong and Zhang, Beichen and Hui, Binyuan and Zheng, Bo and Yu, Bowen and Gao, Chang and Huang, Chengen and Lv, Chenxu and Zheng, Chujie and Liu, Dayiheng and Zhou, Fan and Huang, Fei and Hu, Feng and Ge, Hao and Wei, Haoran and Lin, Huan and Tang, Jialong and Yang, Jian and Tu, Jianhong and Zhang, Jianwei and Yang, Jianxin and Yang, Jiaxi and Zhou, Jing and Zhou, Jingren and Lin, Junyang and Dang, Kai and Bao, Keqin and Yang, Kexin and Yu, Le and Deng, Lianghao and Li, Mei and Xue, Mingfeng and Li, Mingze and Zhang, Pei and Wang, Peng and Zhu, Qin and Men, Rui and Gao, Ruize and Liu, Shixuan and Luo, Shuang and Li, Tianhao and Tang, Tianyi and Yin, Wenbiao and Ren, Xingzhang and Wang, Xinyu and Zhang, Xinyu and Ren, Xuancheng and Fan, Yang and Su, Yang and Zhang, Yichang and Zhang, Yinger and Wan, Yu and Liu, Yuqiong and Wang, Zekun and Cui, Zeyu and Zhang, Zhenru and Zhou, Zhipeng and Qiu, Zihan},
	title = {{{Qwen3} {Technical} {Report}}},
	journal = {arXiv preprint arXiv:2505.09388},
	year = {2025},
	doi = {10.48550/arXiv.2505.09388},
	url = {https://doi.org/10.48550/arXiv.2505.09388},
}

@article{grattafiori_llama_2024,
	author = {Grattafiori, Aaron and Dubey, Abhimanyu and Jauhri, Abhinav and others},
	title = {The {Llama} 3 {Herd} of {Models}},
	journal = {arXiv preprint arXiv:2407.21783},
	year = {2024},
	doi = {10.48550/arXiv.2407.21783},
	url = {https://doi.org/10.48550/arXiv.2407.21783},
}

@techreport{radford_language_nodate,
	title = {Language {Models} are {Unsupervised} {Multitask} {Learners}},
	url = {https://cdn.openai.com/better-language-models/language_models_are_unsupervised_multitask_learners.pdf},
	author = {Radford, Alec and Wu, Jeffrey and Child, Rewon and Luan, David and Amodei, Dario and Sutskever, Ilya},
	institution = {OpenAI},
	year = {2019},
}

@article{deng_implicit_2023,
	author = {Deng, Yuntian and Prasad, Kiran and Fernandez, Roland and Smolensky, Paul and Chaudhary, Vishrav and Shieber, Stuart},
	title = {Implicit {Chain} of {Thought} {Reasoning} via {Knowledge} {Distillation}},
	journal = {arXiv preprint arXiv:2311.01460},
	year = {2023},
	doi = {10.48550/arXiv.2311.01460},
	url = {https://doi.org/10.48550/arXiv.2311.01460},
}

@article{causal_survey,
	author = {Yao, Liuyi and Chu, Zhixuan and Li, Sheng and Li, Yaliang and Gao, Jing and Zhang, Aidong},
	title = {A Survey on Causal Inference},
	journal = {ACM {Transactions} on {Knowledge} {Discovery} from {Data}},
	year = {2021},
	volume = {15},
	number = {5},
	pages = {1--46},
	publisher = {Association for Computing Machinery},
	doi = {10.1145/3444944},
	url = {https://doi.org/10.1145/3444944},
}

@inproceedings{wang-etal-2023-causal,
	author = {Wang, Fei and Mo, Wenjie and Wang, Yiwei and Zhou, Wenxuan and Chen, Muhao},
	title = {A Causal View of Entity Bias in (Large) Language Models},
	booktitle = {Findings of the Association for Computational Linguistics: EMNLP 2023},
	year = {2023},
	month = dec,
	address = {Singapore},
	pages = {15173--15184},
	publisher = {Association for Computational Linguistics},
	doi = {10.18653/v1/2023.findings-emnlp.1013},
	url = {https://aclanthology.org/2023.findings-emnlp.1013/},
}

@inproceedings{yang-etal-2023-causal,
	author = {Yang, Zhen and Liu, Yongbin and Ouyang, Chunping},
	title = {Causal Intervention-based Few-Shot Named Entity Recognition},
	booktitle = {Findings of the Association for Computational Linguistics: EMNLP 2023},
	year = {2023},
	month = dec,
	address = {Singapore},
	pages = {15635--15646},
	publisher = {Association for Computational Linguistics},
	doi = {10.18653/v1/2023.findings-emnlp.1046},
	url = {https://aclanthology.org/2023.findings-emnlp.1046/},
}

@article{kaddour_causal_2022,
	author = {Kaddour, Jean and Lynch, Aengus and Liu, Qi and Kusner, Matt J. and Silva, Ricardo},
	title = {Causal {Machine} {Learning}: {A} {Survey} and {Open} {Problems}},
	journal = {Foundations and {Trends} in {Optimization}},
	year = {2025},
	volume = {9},
	number = {1--2},
	pages = {1--247},
	publisher = {Now Publishers},
	doi = {10.1561/2400000052},
	url = {https://doi.org/10.1561/2400000052},
}

@article{singh_kernel_2022,
	author = {Singh, Rahul and Xu, Liyuan and Gretton, Arthur},
	title = {Kernel {Methods} for {Causal} {Functions}: {Dose}, {Heterogeneous} and {Incremental} {Response} {Curves}},
	journal = {Biometrika},
	year = {2024},
	volume = {111},
	number = {2},
	pages = {497--516},
	publisher = {Oxford University Press},
	doi = {10.1093/biomet/asad042},
	url = {https://doi.org/10.1093/biomet/asad042},
}

@book{dag,
    author = {Thulasiraman, K. and Swamy, M. N. S.},
    title = {Graphs: theory and algorithms},
    year = {1992},
    isbn = {0471513563},
    publisher = {John Wiley \& Sons, Inc.},
    address = {USA}
}

@inproceedings{talmor-etal-2019-commonsenseqa,
	author = {Talmor, Alon and Herzig, Jonathan and Lourie, Nicholas and Berant, Jonathan},
	title = {{C}ommonsense{QA}: A Question Answering Challenge Targeting Commonsense Knowledge},
	booktitle = {Proceedings of the 2019 Conference of the North {A}merican Chapter of the Association for Computational Linguistics: Human Language Technologies, Volume 1 (Long and Short Papers)},
	year = {2019},
	month = jun,
	address = {Minneapolis, Minnesota},
	pages = {4149--4158},
	publisher = {Association for Computational Linguistics},
	doi = {10.18653/v1/N19-1421},
	url = {https://aclanthology.org/N19-1421/},
}

@article{geva-etal-2021-aristotle,
	author = {Geva, Mor and Khashabi, Daniel and Segal, Elad and Khot, Tushar and Roth, Dan and Berant, Jonathan},
	title = {Did {A}ristotle Use a Laptop? A Question Answering Benchmark with Implicit Reasoning Strategies},
	journal = {Transactions of the Association for Computational Linguistics},
	year = {2021},
	volume = {9},
	pages = {346--361},
	publisher = {MIT Press},
	doi = {10.1162/tacl_a_00370},
	url = {https://aclanthology.org/2021.tacl-1.21/},
}

@inproceedings{yu_causal_2025,
	author = {Yu, Xiangning and Wang, Zhuohan and Yang, Linyi and Li, Haoxuan and Liu, Anjie and Xue, Xiao and Wang, Jun and Yang, Mengyue},
	title = {Causal {Sufficiency} and {Necessity} {Improves} {Chain-of-Thought} {Reasoning}},
	booktitle = {Advances in {Neural} {Information} {Processing} {Systems}},
	year = {2025},
	month = dec,
	address = {San Diego, CA, USA and Mexico City, Mexico},
	volume = {38},
    pages = {126109--126141},
	publisher = {Curran Associates, Inc.},
	url = {https://papers.nips.cc/paper_files/paper/2025/file/b7870bd43b2d133a1ed95582ae5d82a4-Paper-Conference.pdf},
}

@article{bai-etal-2025-constparse,
	author = {Bai, Xuefeng and Wu, Jialong and Chen, Yulong and Wang, Zhongqing and Chen, Kehai and Zhang, Min and Zhang, Yue},
	title = {Constituency {Parsing} Using {{LLM}s}},
	journal = {IEEE Transactions on Audio, Speech and Language Processing},
	year = {2025},
	volume = {33},
	pages = {3762--3775},
	doi = {10.1109/TASLPRO.2025.3600867},
}

@inproceedings{rimsky-etal-2024-steering,
	author = {Rimsky, Nina and Gabrieli, Nick and Schulz, Julian and Tong, Meg and Hubinger, Evan and Turner, Alexander},
	title = {Steering {Llama} 2 via {Contrastive} {Activation} {Addition}},
	booktitle = {Proceedings of the 62nd Annual Meeting of the Association for Computational Linguistics (Volume 1: Long Papers)},
	year = {2024},
	month = aug,
	address = {Bangkok, Thailand},
	pages = {15504--15522},
	publisher = {Association for Computational Linguistics},
	doi = {10.18653/v1/2024.acl-long.828},
	url = {https://aclanthology.org/2024.acl-long.828/},
}

@article{zou2025representationengineeringtopdownapproach,
	author = {Andy Zou and Long Phan and Sarah Chen and James Campbell and Phillip Guo and Richard Ren and Alexander Pan and Xuwang Yin and Mantas Mazeika and Ann-Kathrin Dombrowski and Shashwat Goel and Nathaniel Li and Michael J. Byun and Zifan Wang and Alex Mallen and Steven Basart and Sanmi Koyejo and Dawn Song and Matt Fredrikson and J. Zico Kolter and Dan Hendrycks},
	title = {{Representation} {Engineering}: {A} {Top-Down} {Approach} to {{AI}} {Transparency}},
	journal = {arXiv preprint arXiv:2310.01405},
	year = {2023},
	doi = {10.48550/arXiv.2310.01405},
	url = {https://doi.org/10.48550/arXiv.2310.01405},
}

@inproceedings{10.5555/3666122.3667919,
	author = {Li, Kenneth and Patel, Oam and Vi\'{e}gas, Fernanda and Pfister, Hanspeter and Wattenberg, Martin},
	title = {{Inference-Time} {Intervention}: {Eliciting} {Truthful} {Answers} from a {Language} {Model}},
	booktitle = {Advances in {Neural} {Information} {Processing} {Systems}},
	year = {2023},
	month = dec,
	address = {New Orleans, LA, USA},
	volume = {36},
	pages = {41451--41530},
	publisher = {Curran Associates, Inc.},
	url = {https://proceedings.neurips.cc/paper_files/paper/2023/file/81b8390039b7302c909cb769f8b6cd93-Paper-Conference.pdf},
}

@inproceedings{zhang2026evaluatingsteeringmodalitypreferences,
	author = {Yu Zhang and Jinlong Ma and Yongshuai Hou and Xuefeng Bai and Kehai Chen and Yang Xiang and Jun Yu and Min Zhang},
	title = {{Evaluating} and {Steering} {Modality} {Preferences} in {Multimodal} {Large} {Language} {Model}},
	booktitle = {Proceedings of the 43rd {International} {Conference} on {Machine} {Learning}},
	year = {2026},
	month = jul,
	address = {Seoul, South Korea},
	doi = {10.48550/arXiv.2505.20977},
	url = {https://doi.org/10.48550/arXiv.2505.20977},
}

@article{zhu2025scalinglatentreasoninglooped,
	author = {Rui-Jie Zhu and Zixuan Wang and Kai Hua and Tianyu Zhang and Ziniu Li and Haoran Que and Boyi Wei and Zixin Wen and Fan Yin and He Xing and Lu Li and Jiajun Shi and Kaijing Ma and Shanda Li and Taylor Kergan and Andrew Smith and Xingwei Qu and Mude Hui and Bohong Wu and Qiyang Min and Hongzhi Huang and Xun Zhou and Wei Ye and Jiaheng Liu and Jian Yang and Yunfeng Shi and Chenghua Lin and Enduo Zhao and Tianle Cai and Ge Zhang and Wenhao Huang and Yoshua Bengio and Jason Eshraghian},
	title = {{Scaling} {Latent} {Reasoning} via {Looped} {Language} {Models}},
	journal = {arXiv preprint arXiv:2510.25741},
	year = {2025},
	doi = {10.48550/arXiv.2510.25741},
	url = {https://doi.org/10.48550/arXiv.2510.25741},
}

@article{eit2025latentcot,
	author = {Xinghao Chen and Anhao Zhao and Heming Xia and Xuan Lu and Hanlin Wang and Yanjun Chen and Wei Zhang and Jian Wang and Wenjie Li and Xiaoyu Shen},
	title = {{Reasoning} {Beyond} {Language}: {A} {Comprehensive} {Survey} on {Latent} {Chain-of-Thought} {Reasoning}},
	journal = {arXiv preprint arXiv:2505.16782},
	year = {2025},
	doi = {10.48550/arXiv.2505.16782},
	url = {https://doi.org/10.48550/arXiv.2505.16782},
}

@inproceedings{conmy_towards_2023,
	author = {Arthur Conmy and Augustine N. Mavor{-}Parker and Aengus Lynch and Stefan Heimersheim and Adri{\`{a}} Garriga{-}Alonso},
	title = {{Towards} {Automated} {Circuit} {Discovery} for {Mechanistic} {Interpretability}},
	booktitle = {Advances in {Neural} {Information} {Processing} {Systems}},
	year = {2023},
	month = dec,
	address = {New Orleans, LA, USA},
	volume = {36},
	pages = {16318--16352},
	publisher = {Curran Associates, Inc.},
	url = {https://proceedings.neurips.cc/paper_files/paper/2023/file/34e1dbe95d34d7ebaf99b9bcaeb5b2be-Paper-Conference.pdf},
}

@inproceedings{wang_interpretability_2023,
	author = {Wang, Kevin Ro and Variengien, Alexandre and Conmy, Arthur and Shlegeris, Buck and Steinhardt, Jacob},
	title = {{Interpretability} in the {Wild}: {A} {Circuit} for {Indirect} {Object} {Identification} in {{GPT}-2} {Small}},
	booktitle = {The {Eleventh} {International} {Conference} on {Learning} {Representations}},
	year = {2023},
	month = may,
	address = {Kigali, Rwanda},
	url = {https://openreview.net/forum?id=NpsVSN6o4ul},
}

@inproceedings{meng_locating_2022,
	author = {Meng, Kevin and Bau, David and Andonian, Alex and Belinkov, Yonatan},
	title = {Locating and {Editing} {Factual} {Associations} in {GPT}},
	booktitle = {Advances in {Neural} {Information} {Processing} {Systems}},
	year = {2022},
	month = dec,
	address = {New Orleans, LA, USA},
	volume = {35},
	pages = {17359--17372},
	publisher = {Curran Associates, Inc.},
	url = {https://proceedings.neurips.cc/paper_files/paper/2022/file/6f1d43d5a82a37e89b0665b33bf3a182-Paper-Conference.pdf},
}

@inproceedings{stolfo_mechanistic_2023,
	author = {Stolfo, Alessandro and Belinkov, Yonatan and Sachan, Mrinmaya},
	title = {A {Mechanistic} {Interpretation} of {Arithmetic} {Reasoning} in {Language} {Models} using {Causal} {Mediation} {Analysis}},
	booktitle = {Proceedings of the 2023 {Conference} on {Empirical} {Methods} in {Natural} {Language} {Processing}},
	year = {2023},
	month = dec,
	address = {Singapore},
	pages = {7035--7052},
	publisher = {Association for Computational Linguistics},
	doi = {10.18653/v1/2023.emnlp-main.435},
	url = {https://doi.org/10.18653/v1/2023.emnlp-main.435},
}

@inproceedings{finlayson_causal_2021,
	author = {Finlayson, Matthew and Mueller, Aaron and Gehrmann, Sebastian and Shieber, Stuart and Linzen, Tal and Belinkov, Yonatan},
	title = {Causal Analysis of Syntactic Agreement Mechanisms in Neural Language Models},
	booktitle = {Proceedings of the 59th Annual Meeting of the Association for Computational Linguistics and the 11th International Joint Conference on Natural Language Processing (Volume 1: Long Papers)},
	year = {2021},
	month = aug,
	address = {Online},
	pages = {1828--1843},
	publisher = {Association for Computational Linguistics},
	doi = {10.18653/v1/2021.acl-long.144},
	url = {https://aclanthology.org/2021.acl-long.144/},
}

@article{robins_identifiability_1992,
	author = {Robins, James M. and Greenland, Sander},
	title = {Identifiability and {Exchangeability} for {Direct} and {Indirect} {Effects}},
	journal = {Epidemiology},
	year = {1992},
	volume = {3},
	number = {2},
	pages = {143--155},
	url = {https://journals.lww.com/epidem/abstract/1992/03000/identifiability_and_exchangeability_for_direct_and.13.aspx},
}

@article{imai_general_2010,
	author = {Imai, Kosuke and Keele, Luke and Tingley, Dustin},
	title = {A general approach to causal mediation analysis},
	journal = {Psychological Methods},
	year = {2010},
	volume = {15},
	number = {4},
	pages = {309--334},
	publisher = {American Psychological Association},
	doi = {10.1037/a0020761},
}

@article{mueller_quest_2026,
	author = {Mueller, Aaron and Brinkmann, Jannik and Li, Millicent and Marks, Samuel and Pal, Koyena and Prakash, Nikhil and Rager, Can and Sankaranarayanan, Aruna and Sharma, Arnab Sen and Sun, Jiuding and Todd, Eric and Bau, David and Belinkov, Yonatan},
	title = {The {Quest} for the {Right} {Mediator}: {Surveying} {Mechanistic} {Interpretability} for {{NLP}} through the {Lens} of {Causal} {Mediation} {Analysis}},
	journal = {Computational {Linguistics}},
	year = {2026},
	volume = {52},
	number = {1},
	pages = {331--378},
	publisher = {{MIT} Press},
	doi = {10.1162/COLI.a.572},
	url = {https://doi.org/10.1162/COLI.a.572},
}

@inproceedings{vig_investigating_2020,
	author = {Vig, Jesse and Gehrmann, Sebastian and Belinkov, Yonatan and Qian, Sharon and Nevo, Daniel and Singer, Yaron and Shieber, Stuart},
	title = {{Investigating} {Gender} {Bias} in {Language} {Models} {Using} {Causal} {Mediation} {Analysis}},
	booktitle = {Advances in {Neural} {Information} {Processing} {Systems}},
	year = {2020},
	month = dec,
	address = {virtual},
	volume = {33},
	pages = {12388--12401},
	publisher = {Curran Associates, Inc.},
	url = {https://proceedings.neurips.cc/paper_files/paper/2020/file/92650b2e92217715fe312e6fa7b90d82-Paper.pdf},
}

@inproceedings{chrysostomou-aletras-2021-improving,
    title = "Improving the Faithfulness of Attention-based Explanations with Task-specific Information for Text Classification",
    author = "Chrysostomou, George  and
      Aletras, Nikolaos",
    booktitle = "Proceedings of the 59th Annual Meeting of the Association for Computational Linguistics and the 11th International Joint Conference on Natural Language Processing (Volume 1: Long Papers)",
    month = aug,
    year = "2021",
    address = "Online",
    publisher = "Association for Computational Linguistics",
    url = "https://aclanthology.org/2021.acl-long.40/",
    doi = "10.18653/v1/2021.acl-long.40",
    pages = "477--488",
}

@inproceedings{serrano-smith-2019-attention,
    title = "Is Attention Interpretable?",
    author = "Serrano, Sofia  and
      Smith, Noah A.",
    booktitle = "Proceedings of the 57th Annual Meeting of the Association for Computational Linguistics",
    month = jul,
    year = "2019",
    address = "Florence, Italy",
    publisher = "Association for Computational Linguistics",
    url = "https://aclanthology.org/P19-1282/",
    doi = "10.18653/v1/P19-1282",
    pages = "2931--2951",
}

@article{zhang_survey_rl_lrm_2025,
  title = {{A Survey of Reinforcement Learning for Large Reasoning Models}},
  author = {Zhang, Kaiyan and others},
  journal = {arXiv preprint arXiv:2509.08827},
  year = {2025},
  doi = {10.48550/arXiv.2509.08827},
  url = {https://doi.org/10.48550/arXiv.2509.08827}
}

@article{wang_beyond_majority_2025,
  title = {{Beyond Majority Voting: Towards Fine-grained and More Reliable Reward Signal for Test-Time Reinforcement Learning}},
  author = {Wang, Weiqin and Wang, Yile and Chen, Kehao and Huang, Hui},
  journal = {arXiv preprint arXiv:2512.15146},
  year = {2025},
  doi = {10.48550/arXiv.2512.15146},
  url = {https://doi.org/10.48550/arXiv.2512.15146}
}

@article{ruan_critical_token_2025,
  title = {{Enhancing Large Language Model Reasoning via Selective Critical Token Fine-Tuning}},
  author = {Ruan, Zhiwen and Li, Yixia and Zhu, He and Chen, Yun and Li, Peng and Liu, Yang and Chen, Guanhua},
  journal = {arXiv preprint arXiv:2510.10974},
  year = {2025},
  doi = {10.48550/arXiv.2510.10974},
  url = {https://doi.org/10.48550/arXiv.2510.10974}
}

@article{xiao_eagle_2025,
  title = {{Enhancing Uncertainty Estimation in LLMs with Expectation of Aggregated Internal Belief}},
  author = {Xiao, Zeguan and Dou, Diyang and Xiong, Boya and Chen, Yun and Chen, Guanhua},
  journal = {arXiv preprint arXiv:2509.01564},
  year = {2025},
  doi = {10.48550/arXiv.2509.01564},
  url = {https://doi.org/10.48550/arXiv.2509.01564}
}

@article{yan_inftythink_plus_2026,
  title = {{{InftyThink+}: Effective and Efficient Infinite-Horizon Reasoning via Reinforcement Learning}},
  author = {Yan, Yuchen and Jiang, Liang and Jiang, Jin and Li, Shuaicheng and Wen, Zujie and Zhang, Zhiqiang and Zhou, Jun and Shao, Jian and Zhuang, Yueting and Shen, Yongliang},
  journal = {arXiv preprint arXiv:2602.06960},
  year = {2026},
  doi = {10.48550/arXiv.2602.06960},
  url = {https://doi.org/10.48550/arXiv.2602.06960}
}
\bibliographystyle{icml2026}

\newpage
\appendix
\onecolumn
\section{Implementation Details}
\label{app:implementation}

\subsection{Training settings for reproduced \textsc{Coconut} and \textsc{CODI}}
\label{app:training_settings}
We reproduce \textsc{Coconut} across three backbones (GPT-2, Llama3-1B, and Qwen3-4B-Instruct) and three datasets (GSM8K, CommonsenseQA, and StrategyQA) using the standard stage-wise latent-replacement curriculum.
On GSM8K, we follow the official two-stage recipe (\emph{CoT-SFT $\rightarrow$ Coconut}): the model is first trained with explicit CoT supervision, and then continued with Coconut-style latent reasoning initialized from the CoT checkpoint.
For GPT-2, both stages are trained for 25 epochs (25+25). For larger backbones, we use a shorter schedule (5+10) consistent with the large-model setting.
On CommonsenseQA and StrategyQA, we adopt a uniform latent curriculum that progressively increases the latent-step budget up to $T{=}6$ using one epoch per latent stage (6 epochs total), followed by 4 additional epochs at the final stage (i.e., 6+4).
Across backbones, we use AdamW-style optimization with weight decay $0.01$ and a backbone-dependent learning rate: GPT-2 runs use $\mathrm{lr}=10^{-4}$, while Llama/Qwen runs use $\mathrm{lr}=10^{-5}$.

For \textsc{CODI}, we use \emph{officially released} checkpoints for GSM8K on GPT-2 and Llama3-1B.
All other \textsc{CODI} results in this paper are obtained from our reproduced checkpoints trained on the remaining dataset/backbone combinations.
We follow the official CODI training pipeline (LoRA-based fine-tuning with a fixed number of latent tokens, cosine learning-rate scheduling with warmup, mixed-precision training, and the projection head enabled in our runs), while allowing dataset/backbone-dependent learning rates for stable optimization.
Concretely, our reproduced CODI runs use learning rates in the range $[5\times 10^{-6},\,3\times 10^{-3}]$ depending on the backbone and dataset (e.g., $3\times 10^{-3}$ for GPT-2, $8\times 10^{-4}$ for Llama-1B, and $2\times 10^{-4}/10^{-5}/5\times 10^{-6}$ for Qwen3-4B-Instruct on GSM8K-Aug/CommonsenseQA/StrategyQA, respectively).
This yields CODI checkpoints matched to our Coconut reproductions in latent-step budget ($T{=}6$) and task coverage, enabling controlled comparisons under identical intervention and readout protocols.

\subsection{Intervention operators: robustness and choice of zero overwrite}
\label{app:intervention_operators}

Our step-wise necessity analysis (RQ1) instantiates the single-step $\mathrm{do}$-intervention (Def.~\ref{def:do_intervention}) via a concrete \emph{intervention operator} that maps a realized latent state $h_t$ to an edited state $\tilde h_t$.
To verify that our conclusions are not tied to a particular operator, we compare six commonly used perturbations that preserve the same causal interface: overwrite one step, then recompute downstream computation under fixed $x$ and $\theta$.
Because activation-level interventions can be sensitive to the choice of corruption or replacement operator, we treat operator variants as robustness checks rather than as uniquely identified causal estimands. This follows the caution used in activation patching, causal tracing, and mediator-search work, where measured effects can depend on the chosen mediator, intervention operator, and readout protocol \citep{meng_locating_2022,conmy_towards_2023,mueller_quest_2026}.

\begin{table}[h]
\centering
\small
\caption{Intervention operators used in the robustness check. 
Here, $\mu$ denotes the global mean latent state, $\mu_t$ denotes the step-specific mean at step $t$, $\epsilon \sim \mathcal{N}(0,I)$, and $\sigma$ is a fixed noise scale.}
\label{tab:intervention_operators}
\begin{tabular}{lll}
\toprule
\textbf{Operator} & \textbf{Description} & \textbf{Edited state $\tilde h_t$} \\
\midrule
\texttt{zero} 
& Replace latent with zeros 
& $\mathbf{0}$ \\

\texttt{mean} 
& Replace with global mean 
& $\mu$ \\

\texttt{mean\_step} 
& Replace with step-specific mean 
& $\mu_t$ \\

\texttt{gaussian\_h} 
& Add Gaussian noise to $h_t$ 
& $h_t + \sigma\epsilon$ \\

\texttt{gaussian\_mu} 
& Add Gaussian noise around global mean 
& $\mu + \sigma\epsilon$ \\

\texttt{gaussian\_mu\_step} 
& Add Gaussian noise around step-specific mean 
& $\mu_t + \sigma\epsilon$ \\
\bottomrule
\end{tabular}
\end{table}

\begin{figure*}[t]
  \centering
  \begin{minipage}{0.49\textwidth}
    \centering
    \includegraphics[width=\textwidth]{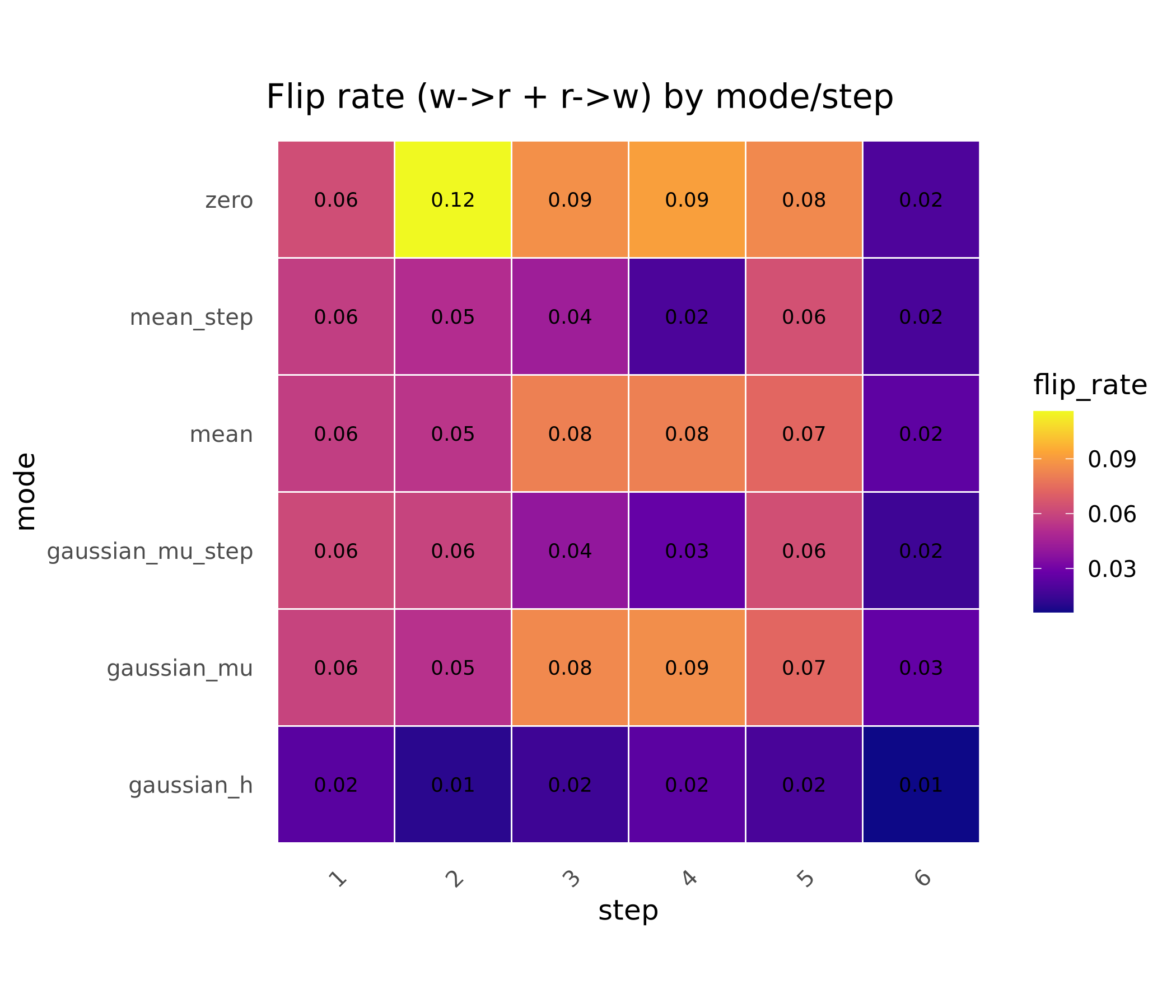}
    \subcaption{\textbf{GSM8K.}}
    \label{fig:app_intervention_gsm8k_codi_llama}
  \end{minipage}\hfill
  \begin{minipage}{0.49\textwidth}
    \centering
    \includegraphics[width=\textwidth]{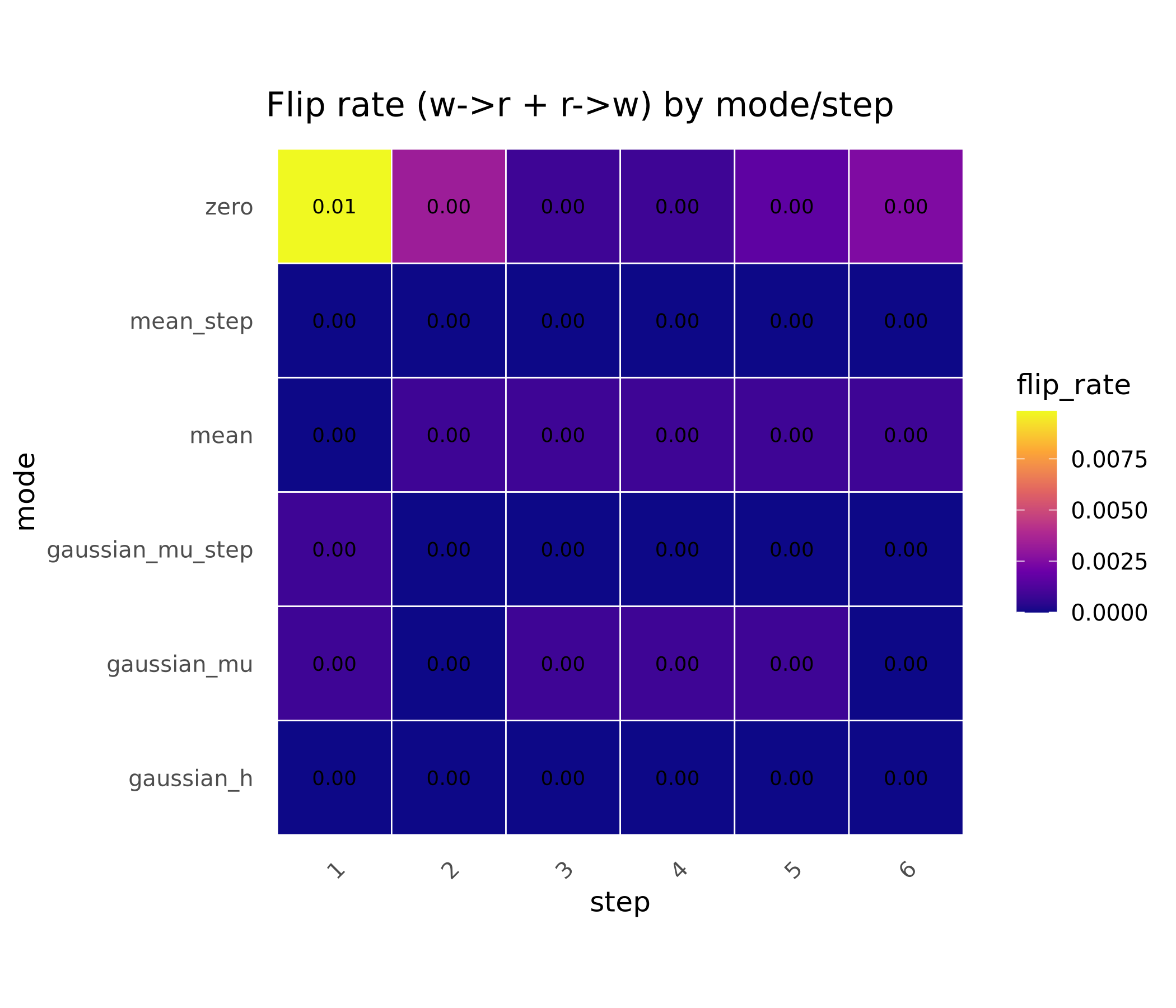}
    \subcaption{\textbf{CommonsenseQA.}}
    \label{fig:app_intervention_csqa_codi_llama}
  \end{minipage}
  \caption{\textbf{Intervention-operator comparison (CODI Llama3-1B).}
  Each cell shows the flip rate $\mathrm{Flip}(t)$, the fraction of examples whose decoded final prediction changes when intervening at step $t$ aggregated over wrong$\rightarrow$right and right$\rightarrow$wrong flips.}
  \label{fig:app_intervention_examples}
\end{figure*}

Figures~\ref{fig:app_intervention_gsm8k_codi_llama} and~\ref{fig:app_intervention_csqa_codi_llama} show two representative examples on \textsc{CODI} (Llama3-1B) for GSM8K and CommonsenseQA, respectively.
Each heatmap cell reports the \textbf{flip rate} $\mathrm{Flip}(t)$ used throughout RQ1, computed as the fraction of examples whose final decoded prediction changes under the intervention at step $t$.
Following our main metric, we aggregate both wrong$\rightarrow$right and right$\rightarrow$wrong flips (i.e., any decision change relative to the baseline rollout), so larger values indicate stronger decision-level dependence on the intervened step.
Across operators, the qualitative step-wise sensitivity patterns are stable: operators that substantially perturb the latent state yield similar relative trends over steps, while weaker/noisier operators typically reduce absolute flip rates without altering the overall step profile.
This robustness indicates that our RQ1 findings are not driven by a specific perturbation choice.

We adopt \textbf{zero overwrite} (\texttt{zero}) as the default operator for the main paper for two practical reasons.
First, it is deterministic and parameter-free, eliminating tuning choices (e.g., the noise scale $\sigma$) and reducing variance across runs.
Second, it applies uniformly across architectures and training recipes, making cross-model comparisons more fair: the intervention strength does not depend on backbone-specific hidden-state norms or distributional statistics beyond the shared representation space.
We therefore use \texttt{zero} throughout the main experiments for reproducibility and interpretability, and treat the remaining operators as sanity checks that validate the stability of our conclusions.

\subsection{Teacher-forced readouts and influence matrix}
\label{app:teacher-forced}

To measure propagation effects with minimal sampling noise, we use teacher-forced readouts on a canonical gold-answer string.
The answer template follows each method's training paradigm (rather than the dataset): for \textsc{Coconut} we use
``[prefix] \texttt{\#\#\#} \{answer\}'', while for \textsc{CODI} we use ``[prefix] The answer is \{answer\}''.
We denote the resulting gold answer token sequence by $a_{1:L}$.

\paragraph{Teacher-forced distributions.}
Given a trajectory (baseline or intervened) and a designated readout step $s$, we compute teacher-forced logits for the gold answer tokens $a_{1:L}$ and obtain token-level predictive distributions
\begin{equation}
p^{(s)}(a_\ell \mid a_{<\ell}, x)\;=\;\mathrm{softmax}\!\big(\mathbf{z}^{(s)}_\ell\big),\qquad \ell=1,\dots,L,
\end{equation}
where $\mathbf{z}^{(s)}_\ell$ denotes the teacher-forced logit vector at gold token position $\ell$ when reading out from step $s$ (with the same $x$ and fixed parameters $\theta$).
We compute these distributions for both the baseline rollout ($p_{\text{base}}^{(s)}$) and the intervened rollout ($p_{\text{int}}^{(s)}$).

\paragraph{Influence matrix from token-averaged KL.}
For an intervention applied at step $t$ and a readout at step $s$, we define the influence weight $W_{t,s}$ (Eq.~\ref{eq:rq2_W}) as the token-averaged KL divergence between the baseline and intervened teacher-forced distributions on the gold answer:
\begin{equation}
W_{t,s}\;=\;\frac{1}{L}\sum_{\ell=1}^{L}
D_{\mathrm{KL}}\!\left(p_{\text{base}}^{(s)}(\cdot \mid a_{<\ell},x)\ \big\|\ p_{\text{int}}^{(s)}(\cdot \mid a_{<\ell},x)\right).
\label{eq:app_teacher_forced_W}
\end{equation}
Intuitively, $W_{t,s}$ measures how much an intervention at step $t$ changes the model's predictive distribution over the gold answer when the trajectory is read out at step $s$.
We aggregate $W_{t,s}$ over examples by averaging across the evaluation set.

\paragraph{Visualization and sparsification.}
To visualize influence graphs, we apply the same sparsification protocol as in the main text.
Specifically, we threshold edges at $\alpha \cdot \max(W)$ with $\alpha=0.1$, and for each source step we retain only the top-1 outgoing edge for readability.
Edge weights in figures correspond to the (dataset-averaged) dense matrix entries $W_{t,s}$ prior to sparsification.

\paragraph{Why teacher forcing.}
Teacher forcing isolates propagation effects from output sampling variability: it evaluates changes in the model's conditional distribution along a fixed gold answer path, rather than changes in a sampled decoded string.
This makes $W$ a more stable proxy for step-to-step influence, especially when the decoded answer is short or when sampling induces high variance across rollouts.

\section{Dataset Information}
\label{app:data}

\subsection{Dataset Statistics}
Table~\ref{tab:dataset_info} reports the dataset split sizes used in our experiments.
All evaluations use the original benchmark test sets from the corresponding dataset papers.
For training, we use CoT-augmented variants when required by the training paradigms: for GSM8K we train on GSM8K-Aug from~\citep{deng_implicit_2023} while evaluating on the original GSM8K test set~\citep{cobbe_training_2021}; for CommonsenseQA and StrategyQA, we use CoT-augmented training data released by \textsc{CODI}~\citep{shen_codi_2025} and evaluate on the original test sets~\citep{talmor-etal-2019-commonsenseqa,geva-etal-2021-aristotle}.
These augmentations affect training-time supervision only; we do not modify any benchmark test set.

\begin{table}[ht]
\caption{\textbf{Dataset statistics used in our experiments.}
Train sizes correspond to the training splits actually used for model training (GSM8K-Aug~\citep{deng_implicit_2023}; CommonsenseQA-CoT and StrategyQA-CoT from \textsc{CODI}~\citep{shen_codi_2025}).
Test sizes correspond to the original benchmark test splits~\citep{talmor-etal-2019-commonsenseqa,cobbe_training_2021,geva-etal-2021-aristotle}.}
\label{tab:dataset_info}
\centering
\begin{small}
\begin{sc}
\begin{tabular}{lrr}
\hline
\textbf{Dataset} & \textbf{Train} & \textbf{Test} \\
\hline
CommonsenseQA & 8{,}096 & 1{,}221 \\
GSM8K & 385{,}620 & 1{,}319 \\
StrategyQA & 1{,}809 & 229 \\
\hline
\end{tabular}
\end{sc}
\end{small}
\end{table}

\subsection{Dataset Examples}
\label{app:data_examples}

The following examples illustrate the CoT-augmented training format used in our runs.
CoT rationales are \emph{not} part of the original benchmarks; they are training-time supervision taken from the corresponding augmented variants (GSM8K-Aug~\citep{deng_implicit_2023} and \textsc{CODI}-released CoT data~\citep{shen_codi_2025}).
We evaluate on the original benchmark test sets without modification.

\begin{databox}[label=box:gsm8k]{GSM8K-Aug (training)}
\textbf{Question:} Out of 600 employees in a company, 30\% got promoted while 10\% received bonus.
How many employees did not get either a promotion or a bonus?

\textbf{CoT:} 600*30/100=180, 600*10/100=60, 600-240=360

\textbf{Answer:} 360
\end{databox}

\begin{databox}[label=box:commonsense]{CommonsenseQA-CoT (training)}
\textbf{Question:} Google Maps and other highway and street GPS services have replaced what?
Choices: A: united states \ \ B: mexico \ \ C: countryside \ \ D: atlas \ \ E: oceans

\textbf{CoT:} Google Maps and other GPS services are tools used for navigation and mapping, which have largely replaced traditional paper maps and atlases that people used in the past for similar purposes.
The other options, such as countries and geographic features, do not relate to navigational tools.
Thus, the best answer is D: atlas.

\textbf{Answer:} D
\end{databox}

\begin{databox}[label=box:strategy]{StrategyQA-CoT (training)}
\textbf{Question:} Would a dog respond to bell before Grey seal?

\textbf{CoT:} To determine if a dog would respond to a bell before a Grey seal, we need to consider the context and conditioning involved.
Dogs are often trained to respond to bells due to their domestication and close relationships with humans.
Classic conditioning experiments, like Pavlov's dogs, indicate that dogs can be trained to associate a bell with specific actions, such as being fed.
Grey seals, on the other hand, are wild marine mammals that are not typically subjected to the same kind of training or conditioning that dogs regularly undergo.
Therefore, without specific conditioning, a Grey seal is unlikely to naturally associate or respond to a bell in the same way a trained dog would.
Thus, a dog, particularly one that has been conditioned, is expected to respond to a bell before a Grey seal.

\textbf{Answer:} true
\end{databox}

\newpage

\section{Additional RQ2 Details and Robustness Checks}
\label{app:rq2}
\label{app:rq2_details}
\label{app:additional_rq2}

\subsection{Dense adjacency matrices for influence graphs}
\label{app:rq2_full_mats}

The main text visualizes sparsified influence graphs for readability (threshold $\alpha{=}0.1$ and top-1 outgoing edge per node).
Here we provide the corresponding dense influence matrices $W$ (Eq.~\ref{eq:rq2_W}) as heatmaps, which are the objects used to compute all structure metrics.

\begin{figure*}[h]
  \centering
  \includegraphics[width=0.96\textwidth]{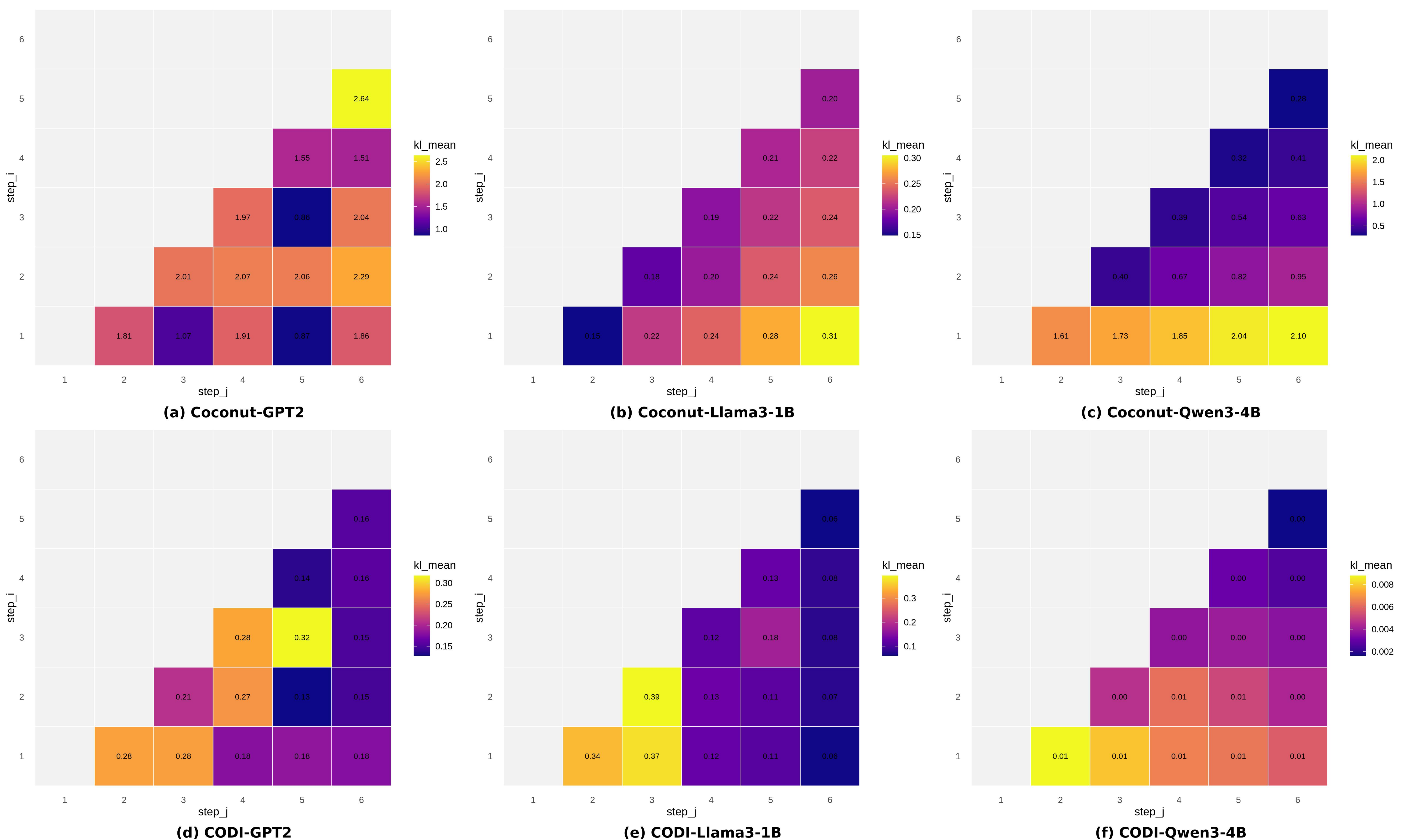}
  \caption{\textbf{Dense influence matrices for latent reasoning (GSM8K).}
  Each cell $(t,s)$ shows $W_{t,s}$ from Eq.~\ref{eq:rq2_W} (teacher-forced KL shift on the gold answer when intervening at $t$ and reading out at $s$).}
  \label{fig:app_rq2_latent_heatmaps_gsm8k}
\end{figure*}

\begin{figure*}[t]
  \centering
  \includegraphics[width=0.96\textwidth]{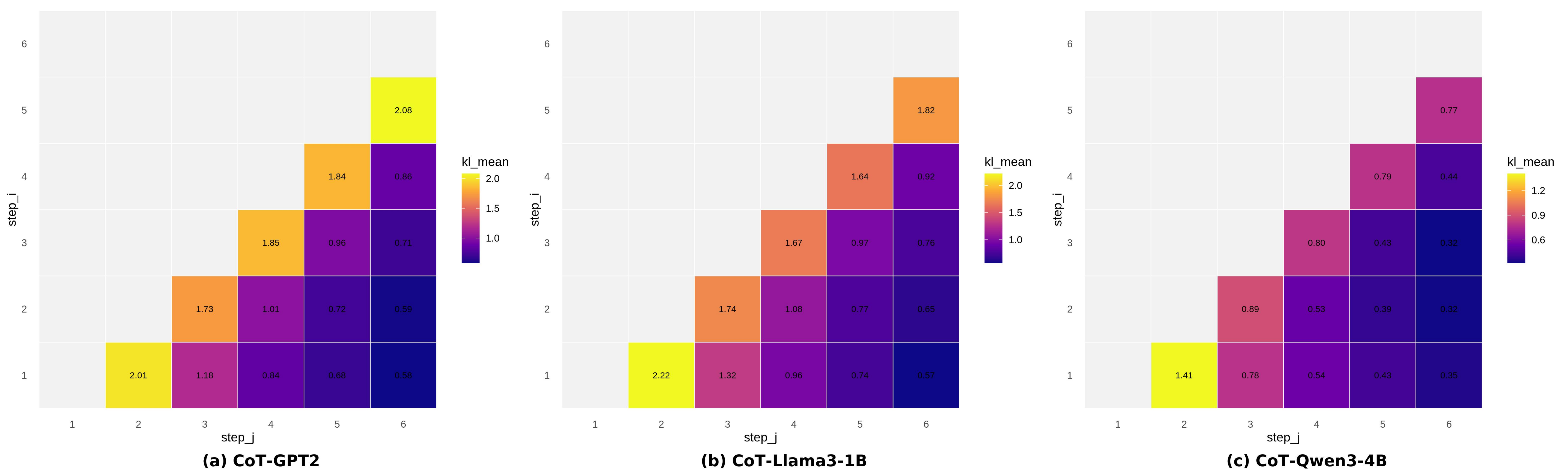}
  \caption{\textbf{Dense influence matrices for explicit CoT (GSM8K; CoT-SFT).}
  Each cell $(t,s)$ shows $W_{t,s}$ from Eq.~\ref{eq:rq2_W} computed on segmented CoT-step states.}
  \label{fig:app_rq2_explicit_heatmaps_gsm8k}
\end{figure*}
\newpage
\subsection{Additional CommonsenseQA results}
\label{app:rq2_csqa}

This subsection provides additional RQ2 results on CommonsenseQA, including dense influence matrices (latent and explicit), sparsified principal influence graphs, and the corresponding structure metrics.
All quantities follow the same construction as in Sec.~\ref{sec:rq2}, based on the dense matrix $W$ from Eq.~\ref{eq:rq2_W}.

\begin{figure*}[h]
  \centering
  \includegraphics[width=0.96\textwidth]{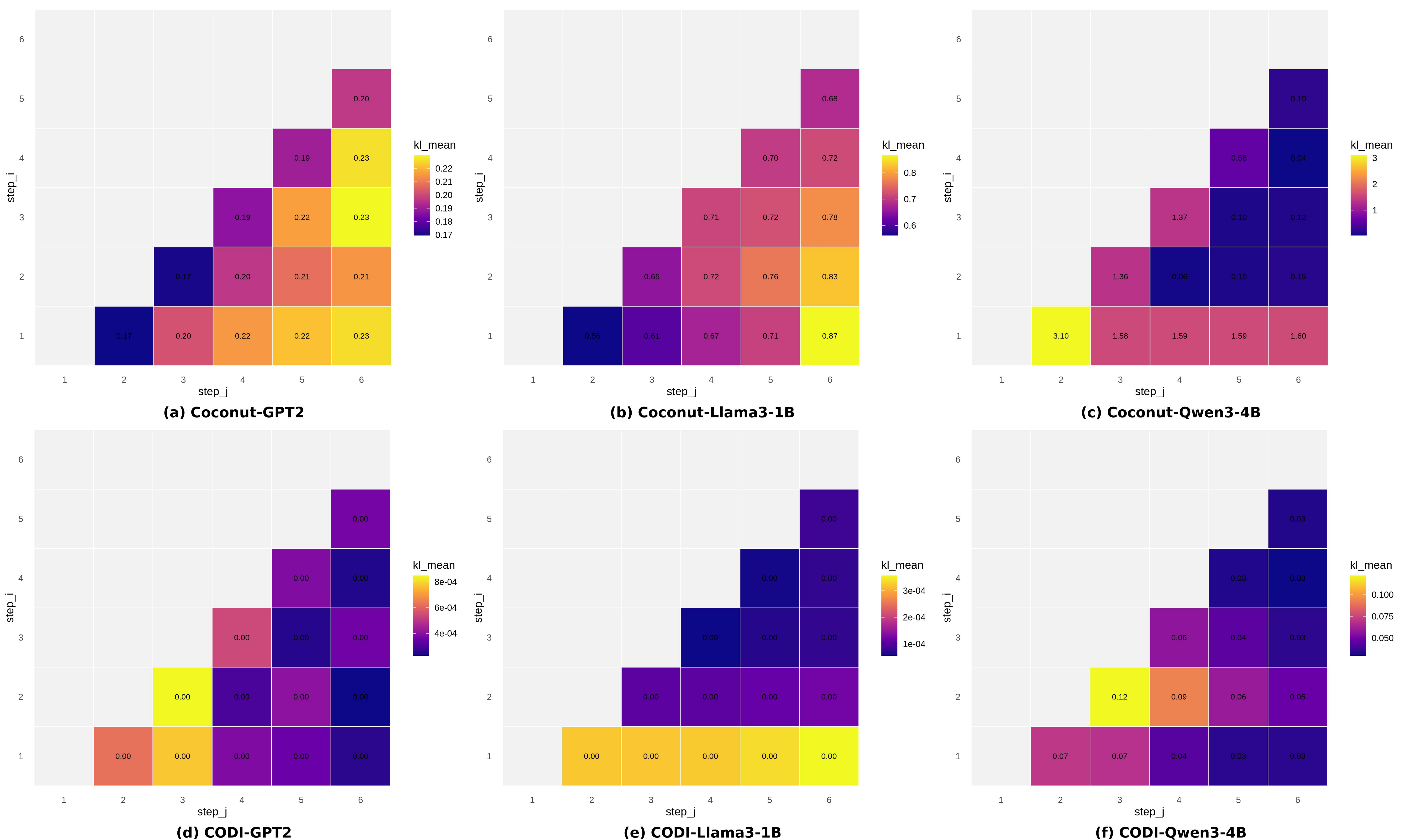}
  \caption{\textbf{Dense influence matrices for latent reasoning (CommonsenseQA; \textsc{Coconut}/\textsc{CODI}).}
  Each cell $(t,s)$ shows $W_{t,s}$ from Eq.~\ref{eq:rq2_W} under teacher-forced readouts.}
  \label{fig:app_rq2_latent_heatmaps_csqa}
\end{figure*}

\begin{figure*}[h]
  \centering
  \includegraphics[width=0.96\textwidth]{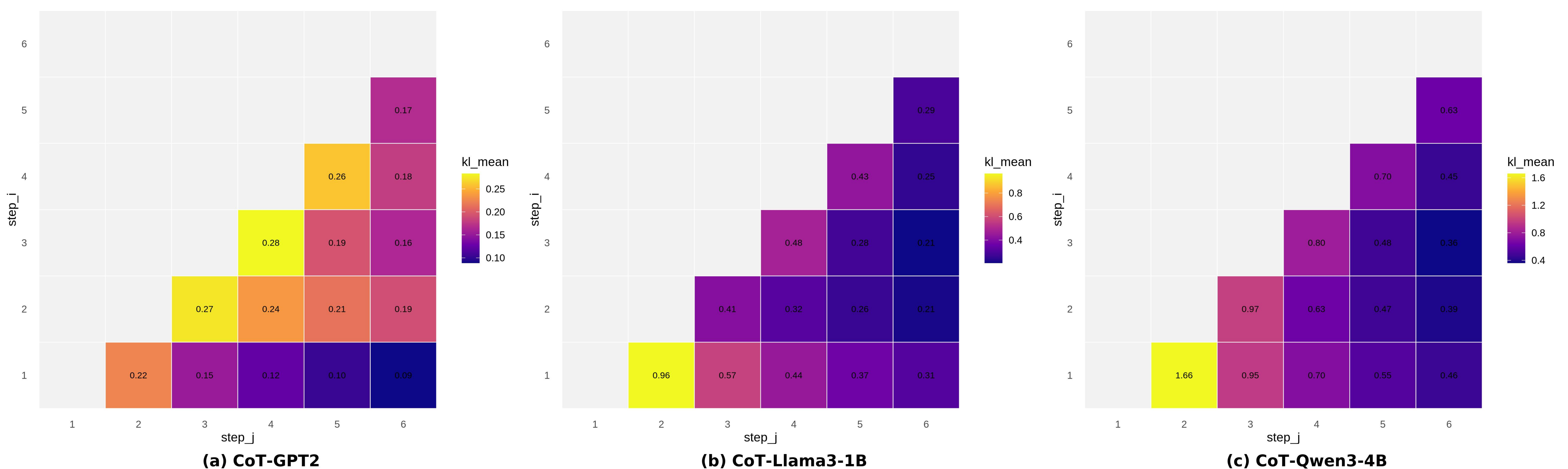}
  \caption{\textbf{Dense influence matrices for explicit CoT (CommonsenseQA; CoT-SFT).}
  Each cell $(t,s)$ shows $W_{t,s}$ from Eq.~\ref{eq:rq2_W} computed on segmented CoT-step states.}
  \label{fig:app_rq2_explicit_heatmaps_csqa}
\end{figure*}

\begin{figure*}[h]
  \centering
  \includegraphics[width=\textwidth]{figs/rq2/commonsenseqa_explicit_graph_grid.pdf}
  \caption{\textbf{Explicit CoT principal influence graphs (CommonsenseQA; CoT-SFT baselines).}
  Nodes denote the first $T{=}6$ segmented CoT steps.
  Edge $t\!\to\!s$ indicates propagation strength $W_{t,s}$ from Eq.~\ref{eq:rq2_W}.
  For readability, we show only top-1 outgoing edges after thresholding at $\alpha{=}0.1\cdot\max(W)$.}
  \label{fig:app_rq2_explicit_graphs_csqa}
\end{figure*}

\begin{figure*}[h]
  \centering
  \includegraphics[width=\textwidth]{figs/rq2/latent_graph_grid_commonsenseqa.pdf}
  \caption{\textbf{Latent principal influence graphs (CommonsenseQA; \textsc{Coconut}/\textsc{CODI}).}
  Nodes are latent steps $t\in\{1,\dots,6\}$.
  Edge weights follow Eq.~\ref{eq:rq2_W} under single-step interventions, rendered with the same sparsification protocol as Figure~\ref{fig:app_rq2_explicit_graphs_csqa}.}
  \label{fig:app_rq2_latent_graphs_csqa}
\end{figure*}

\begin{figure*}[h]
  \centering
  \includegraphics[width=\textwidth]{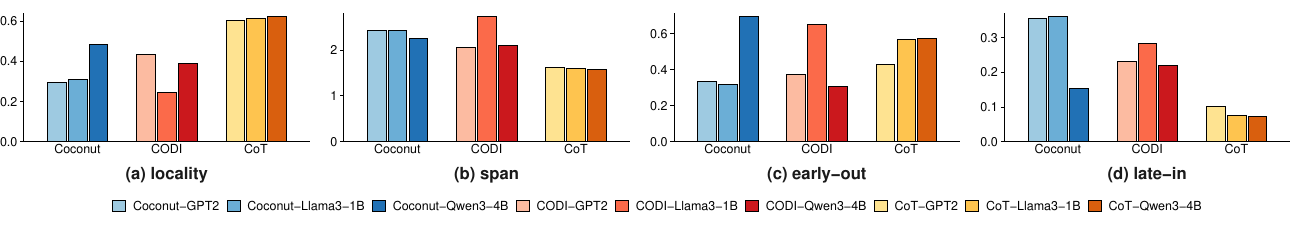}
  \caption{\textbf{Structure metrics on influence graphs (CommonsenseQA).}
  Metrics are computed on the normalized matrix $\bar W$ (Eq.~\ref{eq:app_rq2_normW}) and use the same hyperparameters as the main text.}
  \label{fig:app_rq2_structure_metrics_csqa}
\end{figure*}

\FloatBarrier
\subsection{Definitions of structure metrics}
\label{app:rq2_metrics}

We compute structure metrics on a normalized influence matrix $\bar W$ obtained by re-scaling $W$ over valid entries $(t<s)$:
\begin{equation}
\label{eq:app_rq2_normW}
\bar W_{t,s}=\frac{W_{t,s}}{\sum_{a<b}W_{a,b}+\epsilon}.
\end{equation}
Let $\mathcal{E}=\{(t,s): 1\le t<s\le T\}$.
We define:
\begin{align}
\mathrm{Locality}(k) &= \sum_{(t,s)\in\mathcal{E}}\mathbf{1}\{s-t\le k\}\,\bar W_{t,s},\\
\mathrm{Span} &= \sum_{(t,s)\in\mathcal{E}}(s-t)\,\bar W_{t,s},\\
\mathrm{EarlyOut}(m) &= \sum_{t\le m}\sum_{s>t}\bar W_{t,s},\\
\mathrm{LateIn}(m) &= \sum_{s\ge m}\sum_{t<s}\bar W_{t,s}.
\end{align}
In all experiments we use $k{=}1$ and $m{=}2$ (early) / $m{=}5$ (late) for $T{=}6$.

\subsection{Influence structures under alternative intervention operators}
\label{app:rq2_operator_robustness}

Appendix~\ref{app:intervention_operators} verifies that the RQ1 necessity profiles are qualitatively stable across six intervention operators. We additionally test whether the RQ2 influence structure is specific to zero overwrite. We recompute the influence matrix $W$ from Eq.~\ref{eq:rq2_W} on GSM8K using two non-zero operators: global-mean replacement (\texttt{mean}) and Gaussian perturbation around the realized latent state (\texttt{gaussian\_h}). Both operators preserve the same causal interface as the main experiment: exactly one latent step is modified, downstream computation is recomputed, and propagation is measured by teacher-forced KL at downstream readout steps.

Figure~\ref{fig:app_rq2_operator_robustness} shows the resulting principal influence graphs for \textsc{Coconut}-GPT2 and \textsc{CODI}-GPT2. The absolute edge magnitudes vary with the intervention strength, but the qualitative conclusion is unchanged: influence is non-uniform across steps and is not restricted to adjacent transitions. In particular, non-local routes remain visible under both global-mean replacement and Gaussian perturbation, indicating that the skip-dominant RQ2 pattern is not an artifact of zero overwrite alone.

\begin{figure*}[h]
  \centering
  \includegraphics[width=0.92\textwidth]{figs/rebuttal/rq2_operator_robustness_gpt2.pdf}
  \caption{\textbf{RQ2 operator robustness on GSM8K.}
  We recompute principal influence graphs for \textsc{Coconut}-GPT2 and \textsc{CODI}-GPT2 using global-mean replacement (\texttt{mean}) and Gaussian perturbation around the realized latent state (\texttt{gaussian\_h}). Edge weights are computed with the same teacher-forced KL readout as Eq.~\ref{eq:rq2_W}. Non-local influence remains visible under both operators.}
  \label{fig:app_rq2_operator_robustness}
\end{figure*}

\subsection{Readout-conditioned influence structures}
\label{app:rq2_logit_readout}

The main RQ2 experiment uses answer-level teacher-forced KL because it directly measures how an upstream intervention changes the model's distribution over the gold answer path. To check whether the influence graph depends on this specific readout, we also evaluate a hidden-state LM-head readout that does not condition on the gold answer string. For an intervention at step $t$ and a downstream readout step $s$, we project the baseline and intervened hidden states at step $s$ through the language-model head and compute
\begin{equation}
W^{\mathrm{LM}}_{t,s}
=\mathbb{E}_i\left[
D_{\mathrm{KL}}\!\left(
\mathrm{softmax}(\mathrm{LMHead}(h_{i,s}))
\,\middle\|\,
\mathrm{softmax}(\mathrm{LMHead}(\tilde h^{\mathrm{do}(t)}_{i,s}))
\right)
\right],\quad t<s.
\label{eq:app_logit_ht_W}
\end{equation}
This readout asks whether the perturbation changes the one-step vocabulary distribution directly induced by the downstream latent state, rather than the teacher-forced distribution along a fixed answer template.

Figure~\ref{fig:app_rq2_logit_readout} reports the resulting GSM8K influence graphs. The graphs remain non-uniform, and several models still exhibit long-range cross-step influence. This supports the interpretation of RQ2 as a protocol-conditioned empirical influence structure, while showing that the main non-locality finding is not tied to a single answer-template readout.

\begin{figure*}[h]
  \centering
  \includegraphics[width=0.92\textwidth]{figs/rebuttal/rq2_logit_readout_gsm8k.pdf}
  \caption{\textbf{RQ2 readout robustness on GSM8K.}
  Influence graphs computed with the hidden-state LM-head readout in Eq.~\ref{eq:app_logit_ht_W}, rather than answer-level teacher-forced KL. The main qualitative pattern remains: influence is non-uniform and often non-local across latent steps.}
  \label{fig:app_rq2_logit_readout}
\end{figure*}

\section{Additional Paradigm Analysis}
\label{app:additional_paradigm}
\label{app:simcot}

To test whether the observations are limited to \textsc{Coconut} and \textsc{CODI}, we additionally evaluate \textsc{Sim-CoT}~\citep{wei_sim-cot_2025} on GSM8K using the same $T{=}6$ step interface where available. This experiment is intended as a paradigm-coverage check, not as a claim that the three paradigms exhaust latent reasoning methods.

\begin{figure*}[h]
  \centering
  \includegraphics[width=0.66\textwidth]{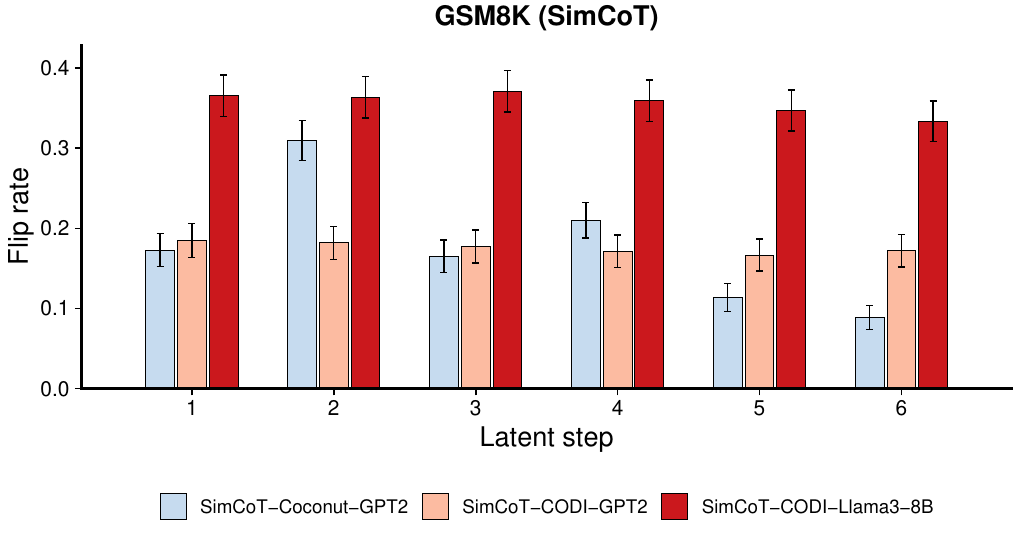}
  \caption{\textbf{RQ1 necessity check for \textsc{Sim-CoT} on GSM8K.}
  Flip rates are computed under the same single-step intervention protocol as Figure~\ref{fig:rq1_flip_overview}. Step importance remains heterogeneous under an additional latent-reasoning paradigm.}
  \label{fig:app_simcot_rq1}
\end{figure*}

Figure~\ref{fig:app_simcot_rq1} shows the RQ1 flip-rate profiles. The Sim-CoT variants also show heterogeneous step importance: the flip rate is not flat across latent steps, and some steps are substantially more decision-relevant than others.

Figure~\ref{fig:app_simcot_rq2} reports the corresponding RQ2 principal influence graphs. Compared with the skip-heavy patterns observed for several \textsc{Coconut}/\textsc{CODI} models, Sim-CoT exhibits relatively more local connectivity in these runs. This difference is consistent with its step-aligned supervision objective: stronger step-level alignment can make adjacent latent transitions more prominent, even though the overall influence structure remains non-uniform.

\begin{figure*}[t]
  \centering
  \includegraphics[width=0.92\textwidth]{figs/rebuttal/rq2_simcot_gsm8k.pdf}
  \caption{\textbf{RQ2 influence graphs for \textsc{Sim-CoT} on GSM8K.}
  The same principal-graph rendering as Figure~\ref{fig:rq2_latent_graphs} is used. Sim-CoT shows more local connectivity than the strongest skip-heavy \textsc{Coconut}/\textsc{CODI} cases, while still displaying non-uniform step-to-step influence.}
  \label{fig:app_simcot_rq2}
\end{figure*}

\FloatBarrier
\section{Additional Model Size Analysis}
\label{app:additional_model_size}
\label{app:coconut_backbone_scaling}

We further replicate the RQ1 and RQ2 analyses on additional \textsc{Coconut} backbones, including Qwen3-1.7B, Mistral-7B, DeepSeek-R1-Qwen2.5-1.5B, and Llama3-8B. These experiments are intended as a qualitative model-size/backbone check rather than a full scaling-law study, because the backbones differ in pretraining, instruction tuning, and reproduction details.

\begin{figure*}[h]
  \centering
  \includegraphics[width=0.66\textwidth]{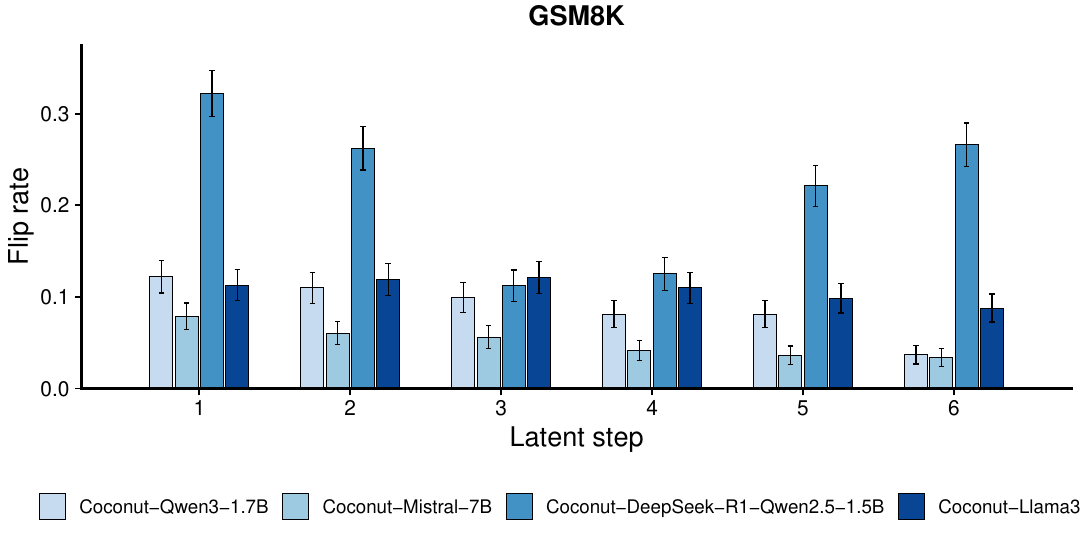}
  \caption{\textbf{RQ1 necessity check on additional \textsc{Coconut} backbones (GSM8K).}
  The magnitude of decision flips differs across backbones, but step-wise sensitivity remains heterogeneous.}
  \label{fig:app_coconut_scaling_rq1}
\end{figure*}

Figure~\ref{fig:app_coconut_scaling_rq1} shows that step-wise flip rates remain heterogeneous across the additional backbones. The absolute flip rate varies substantially by model, but the profiles are not uniformly flat over steps. This supports the main RQ1 claim that latent steps have unequal causal leverage.

Figure~\ref{fig:app_coconut_scaling_rq2} reports the corresponding RQ2 influence graphs. The additional models again show non-uniform propagation, with several long-range edges and model-specific routing patterns. Thus, the main influence-structure observation is not confined to GPT-2/Llama3-1B/Qwen3-4B-Instruct settings, although a systematic scaling study over matched model families remains future work.

\begin{figure*}[h]
  \centering
  \includegraphics[width=0.92\textwidth]{figs/rebuttal/rq2_scaling_coconut_gsm8k.pdf}
  \caption{\textbf{RQ2 influence graphs on additional \textsc{Coconut} backbones (GSM8K).}
  The graphs show non-uniform propagation and model-specific long-range routing across larger and alternative backbones.}
  \label{fig:app_coconut_scaling_rq2}
\end{figure*}

\FloatBarrier
\section{Additional RQ3 Details}
\label{app:rq3}

\subsection{Trajectory sampling and latent-state collection}
\label{app:rq3_stats}

RQ3 studies multi-mode latent dynamics by analyzing multiple trajectories per input.
For each example $x$, we generate $N$ independent rollouts under the same decoding configuration as in the main experiments (temperature/top-$p$/top-$k$ when sampling; greedy decoding when deterministic rollouts are required).
For each rollout, we record (i) the final decoded answer $\hat y$ and (ii) the realized latent states $h_{1:T}=(h_1,\ldots,h_T)$ at the model-defined latent steps, where each $h_t\in\mathbb{R}^d$ is the last-layer hidden state associated with step $t$.

\paragraph{Categorizing answers and defining modes.}
We categorize final answers into a normalized form (e.g., extracting the final option letter for multiple-choice tasks, or applying the same numeric/boolean normalization used for evaluation).
Given the set of normalized answers from $N$ rollouts for the same input, we define the two dominant modes $A$ and $B$ as the two most frequent terminal answers.
Rollouts whose terminal answers fall outside $\{A,B\}$ are treated as residual modes and are excluded from the binary-mode analysis for that input.
This procedure yields a set of labeled latent trajectories $\{(h_{1:T}^{(i)}, m^{(i)})\}_{i=1}^{M}$ with $m^{(i)}\in\{A,B\}$, where $M\le N$ after filtering.

\subsection{Intermediate-step readout implementation details}
\label{app:rq3_readout}

\paragraph{Probes for step-wise mode tracking.}

To measure how mode information evolves across latent steps, we train lightweight probes on frozen latent states.
For each step $t$, we fit a classifier $\pi_{\phi,t}$ that maps the realized latent state $h_t$ to a distribution over modes:
\begin{equation}
\pi_{\phi,t}:\ \mathbb{R}^d \rightarrow \Delta^{2},\qquad
(\hat s_A(t),\hat s_B(t)) = \pi_{\phi,t}(h_t),
\end{equation}
where $\hat s_A(t)$ and $\hat s_B(t)$ denote the probe-predicted probabilities of modes $A$ and $B$ at step $t$.
Unless otherwise stated, we use linear probes (logistic regression) trained with cross-entropy loss and standard $\ell_2$ regularization, which keeps probe capacity minimal and reduces the risk of overfitting artifacts unrelated to the model's latent dynamics.
When class imbalance arises after mode filtering, we balance the probe training set by subsampling to equalize the number of trajectories per mode.

\paragraph{Teacher-forced readouts used in RQ3.}
\label{app:rq3_teacher_forcing}

In addition to probe-based tracking, we use teacher-forced readouts to deterministically score candidate answers and reduce sampling noise when needed.
We follow the same teacher-forcing protocol described in Appendix~\ref{app:teacher-forced} (Teacher-forced readouts).
Importantly, the canonical answer template is method-dependent rather than dataset-dependent:
for \textsc{Coconut} we use ``[prefix] \texttt{\#\#\# \{answer\}}'', while for \textsc{CODI} we use ``[prefix] The answer is \{answer\}''.
Given a readout step $t$, teacher forcing provides a deterministic score (e.g., token-aggregated log-probability) for each candidate answer string, which we use as a complementary signal to the probe outputs in robustness checks.

\subsection{Supporting RQ3 analysis on GSM8K}
\label{app:additional_rq3}
\label{app:rq3_gsm8k}

The main RQ3 analysis uses StrategyQA because its binary answer space provides a clean setting for comparing two answer modes. We additionally run a supporting analysis on GSM8K by retaining prompts for which stochastic rollouts yield two sufficiently frequent dominant numeric answers. Because GSM8K has an open-ended answer space, this filtering is stricter and the retained set is smaller; we therefore use the result as supporting evidence rather than as the central RQ3 experiment.

\begin{figure*}[h]
  \centering
  \includegraphics[width=0.78\textwidth]{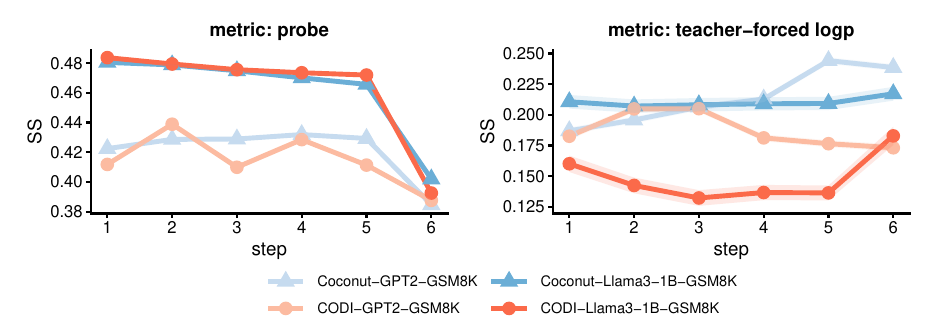}
  \caption{\textbf{Supporting RQ3 analysis on GSM8K.}
  Probe-based and teacher-forced readouts are computed over prompts with two dominant sampled numeric answers. The result supports the distinction between output-level preference and representational commitment, but is less central than StrategyQA due to GSM8K's open-ended answer space.}
  \label{fig:app_rq3_gsm8k}
\end{figure*}

Figure~\ref{fig:app_rq3_gsm8k} shows that the probe-based score remains relatively stable through early and middle latent steps and then drops at the final step, while the teacher-forced log-probability readout is more model-dependent. This is broadly consistent with the main RQ3 interpretation: output-level answer preference and representational commitment need not collapse at the same step, but the GSM8K result should be read cautiously because of the harder two-mode filtering.

\end{document}